\documentclass[lettersize,journal]{IEEEtran}
\usepackage{amsmath,amsfonts,amssymb}
\usepackage{algorithmicx}
\usepackage{algorithm}
\usepackage{algpseudocode}
\usepackage{array}
\usepackage[caption=false,font=normalsize,labelfont=sf,textfont=sf]{subfig}
\usepackage{textcomp}
\usepackage{stfloats}
\usepackage{url}
\usepackage{verbatim}
\usepackage{graphicx}
\usepackage{cite}
\usepackage{xspace}
\usepackage{mathtools}
\usepackage{bm}
\usepackage{tabularx}
\usepackage[table,xcdraw]{xcolor}
\usepackage{booktabs}
\usepackage{multirow}
\usepackage{listings}
\usepackage{enumitem}
\usepackage{soul}
\usepackage[pagebackref=false,breaklinks=true,letterpaper=true,colorlinks,bookmarks=false]{hyperref}
\hypersetup{
    colorlinks=true,
    linkcolor=blue,
    filecolor=magenta,      
}
\usepackage[colorinlistoftodos,prependcaption,textsize=tiny]{todonotes}

\def\pluto{\textsc{Pluto}}
\renewcommand{\paragraph}[1]{\vspace{1.25mm}\noindent\textbf{#1}}

\definecolor{baselinecolor}{gray}{.9}
\newcommand{\baseline}[1]{\cellcolor{baselinecolor}{#1}}

\newcolumntype{x}[1]{>{\centering\arraybackslash}p{#1pt}}
\newcolumntype{y}[1]{>{\raggedright\arraybackslash}p{#1pt}}
\newcolumntype{z}[1]{>{\raggedleft\arraybackslash}p{#1pt}}
\newlength\savewidth

\renewcommand{\paragraph}[1]{\vspace{1.25mm}\noindent\textbf{#1}}

\newcommand{\norm}[1]{\left\lVert#1\right\rVert}
\newcommand{\ulcolor}[2][Red]{\setulcolor{#1}\ul{#2}}

\makeatletter
\DeclareRobustCommand\onedot{\futurelet\@let@token\@onedot}
\def\@onedot{\ifx\@let@token.\else.\null\fi\xspace}
\def\eg{\emph{e.g}\onedot} 
\def\ie{\emph{i.e}\onedot}

\makeatother

\begin{document}

\title{\pluto{}: Pushing the Limit of Imitation Learning-based Planning for Autonomous Driving}

\author{Jie Cheng, Yingbing Chen, and Qifeng Chen
\thanks{Jie Cheng is with the Department of Electronic and Computer Engineering, the Hong Kong University of Science and Technology, Hong Kong SAR (E-mail: \texttt{jchengai@connect.ust.hk})}
\thanks{Yingbing Chen is with Division of Emerging Interdisciplinary Areas, the Hong Kong University of Science and Technology, Hong Kong SAR (E-mail: \texttt{ychengz@connect.ust.hk}).}
\thanks{Qifeng Chen is with the Department of Computer Science and Engineering, the Hong Kong University of Science and Technology, Hong Kong SAR (E-mail: \texttt{cqf@ust.hk}).}%
}

\markboth{}%
{Shell \MakeLowercase{\textit{et al.}}: A Sample Article Using IEEEtran.cls for IEEE Journals}


\maketitle

\begin{abstract}
We present \pluto{}, a powerful framework that \underline{P}ushes the \underline{L}imit of imitation learning-based planning for a\underline{UTO}nomous driving. 
Our improvements stem from three pivotal aspects: 
a longitudinal-lateral aware model architecture that enables flexible and diverse driving behaviors; An innovative auxiliary loss computation method that is broadly applicable and efficient for batch-wise calculation; A novel training framework that leverages contrastive learning, augmented by a suite of new data augmentations to regulate driving behaviors and facilitate the understanding of underlying interactions. 
We assessed our framework using the large-scale real-world nuPlan dataset and its associated standardized planning benchmark.
Impressively, \pluto{} achieves state-of-the-art closed-loop performance, beating other competing learning-based methods and surpassing the current top-performed rule-based planner for the first time. 
Results and code are available at \url{https://jchengai.github.io/pluto}. 
\end{abstract}



\section{Introduction}

\IEEEPARstart{L}{earning}-based planning has emerged as a potentially scalable approach for autonomous driving, attracting significant research interest~\cite{teng2023motion}. Imitation-based planning, in particular, has demonstrated noteworthy success in simulations and real-world applications. 
Yet, the efficacy of learning-based planning remains unsatisfactory. 
As indicated in~\cite{Dauner2023CORL}, conventional rule-based planning outperforms all learning-based alternatives, winning the 2023 nuPlan planning challenge. 
This paper delineates the principal challenges inherent in learning-based planning and presents our novel solutions, aimed at pushing the boundaries of what is achievable with learning-based planning. 

The first challenge lies in acquiring multi-modal driving behaviors. 
It is observed that while learning-based planners are good at learning longitudinal tasks such as lane following, they struggle with lateral tasks~\cite{cheng2023plantf}, for instance, executing lane changes or navigating around obstacles, even when space permits.  
We attribute this deficiency to the absence of explicit lateral behavior modeling within the architectural design of the model. 
Our previous work~\cite{cheng2022mpnp} attempted to address this issue by generating plans that are explicitly conditioned on nearby reference lines, though it was restricted to producing a maximum of three proposals and did not effectively integrate lateral and longitudinal behavior modeling. 
In the present study, we enhance this approach through the adoption of a query-based architecture capable of generating an extensive array of proposals by fusing longitudinal and lateral queries. 
The advanced model design enables our planner to demonstrate diverse and flexible driving behaviors, which we believe is a crucial step toward practical learning-based planning. 

Beyond the model architecture, it is recognized that pure imitation learning encompasses inherent limitations, including the propensity for learning shortcuts~\cite{bansal2018chauffeurnet, muller2005offroad, wang2019monocular,wen2020fighting}, distribution shift~\cite{cheng2023plantf,bansal2018chauffeurnet,zhou2021exploring}, and causal confusion~\cite{de2019causal,cultrera2023addressing} issues. This paper addresses these pervasive challenges across three dimensions:

(1) \textit{Learning beyond pure imitation loss}. We concur with previous studies~\cite{bansal2018chauffeurnet,zhou2021exploring,lu2022imitation_not_enough} that solely relying on imitation loss is insufficient for learning desired driving behaviors. 
It is imperative to impose explicit constraints during the training phase, particularly within the safety-critical realm of autonomous driving.  
A prevalent approach is to add auxiliary losses to penalize adverse behaviors, such as collisions and off-road driving, as previously demonstrated in ~\cite{bansal2018chauffeurnet,zhou2021exploring}. 
However, their methods are either designed for heatmap-based output~\cite{bansal2018chauffeurnet} or need a differentiable rasterizer~\cite{zhou2021exploring} that renders each trajectory point into an image. 
As a result, the output resolution is restricted to reduce the computation burden. 
It remains unclear how to realize these losses efficiently for the more modern vector-based models. 
To bridge this gap, we introduce a novel auxiliary loss calculation methodology predicated on differentiable interpolation. This method not only spans a broad spectrum of auxiliary tasks but also facilitates batch-wise computation within modern deep-learning frameworks, thereby enhancing its applicability and efficiency.

(2) \textit{New data augmentations}. 
Issues arise when the model undergoes open-loop training and closed-loop testing. For instance, the accumulation of errors over time may lead to input data deviating from the training distribution; the model may rely on unintended shortcuts rather than acquiring knowledge. 
Data augmentations have been extensively employed to alleviate these problems and have demonstrated effectiveness, \eg, perturbation-based augmentations~\cite{bansal2018chauffeurnet,zhou2021exploring,cheng2023plantf} teach the model to learn to recover from small deviations and dropout-based augmentations prevent learning shortcuts~\cite{cheng2023plantf}. 
In addition to these two augmentations, we introduce further augmentation techniques aimed at regulating driving behavior and enhancing interaction learning.

(3) \textit{Learning by constrast}.
Imitation learning-based models often struggle to recognize underlying causal relationships due to the absence of interactive feedback with the environment~\cite{de2019causal}. This issue can significantly hinder performance; for instance, a planner may decelerate by mimicking the behavior of nearby agents rather than responding to a red light. Our goal is to address this issue without substantially complicating the training process, as would be the case with reinforcement learning or employing a data-driven simulator. 
Drawing inspiration from the effectiveness of contrastive learning~\cite{chen2020simple}, which enhances representation by differentiating between similar and dissimilar examples, we recognize an opportunity to infuse the model with causal understanding. This is achieved by enabling the model to differentiate between original and modified input data—for example, by excluding the leading vehicles from the autonomous vehicle's (AV's) perspective. Building on this approach and the two previously mentioned strategies, we introduce a novel unified framework termed Contrastive Imitation Learning (CIL).

In summary, this study introduces a comprehensive, data-driven planning framework named \pluto{}, designed to \underline{P}ush the \underline{L}imit of imitation learning-based planning for a\underline{UTO}nomous driving. 
\pluto{} incorporates innovative solutions in model architecture, data augmentation, and the learning framework. 
It has been evaluated using the large-scale real-world nuPlan~\cite{caesar2021nuplan} dataset, where it has demonstrated superior closed-loop performance. 
Notably, \pluto{} surpasses the existing state-of-the-art rule-based planner PDM~\cite{Dauner2023CORL} for the first time, marking a significant milestone in the field. 
Our main contributions are:
\begin{itemize}[leftmargin=*]
    \item We introduce a query-based model architecture that simultaneously addresses lateral and longitudinal planning maneuvers, enabling flexible and diverse driving behaviors. 
    \item We propose a novel method for calculating auxiliary loss based on differential interpolation. This method is applicable to a broad spectrum of auxiliary tasks and allows for efficient batch-wise computation in vector-based models. 
    \item We present the Contrastive Imitation Learning (CIL) framework, accompanied by a new set of data augmentations. The CIL framework is aimed at regulating driving behaviors and enhancing interaction learning, without significantly increasing the complexity of training.
    \item Our evaluation on the large-scale nuPlan dataset demonstrates that \pluto{} achieves state-of-the-art performance in closed-loop planning. Our model and benchmark are publicly available. 
\end{itemize}


\section{Related Work}

\subsection{Imitation-based Planning}
Learning to drive by cloning the policies of experienced drivers is likely the most direct and scalable solution to autonomous driving, considering the abundance and affordability of data today. 
One of the popular methods is end-to-end (E2E) driving~\cite{chen2023e2e}. 
This approach directly learns driving policies from raw sensor data and has made significant strides in a relatively short period. Initially, the focus was on convolutional neural network (CNN)-based models~\cite{chen2020lbc,codevilla2019cilrs} that mapped camera inputs to control policies. This evolved to incorporate more sophisticated methods~\cite{chitta2022transfuser,chitta2021neat,shao2023interfuser,jia2023think} that utilized multi-sensor fusion. More recently, developments spearheaded by entities such as LAV~\cite{chen2022lav} and UniAD~\cite{hu2023planning} have shifted towards a module-based E2E architecture. This approach integrates the processes of perception, prediction, and planning within a unified model~\cite{jiang2023vad, hu2022st-p3}. Despite their potential, most E2E strategies extensively depend on high-fidelity simulation environments like CARLA~\cite{dosovitskiy2017carla} for both training and evaluation. Consequently, these methodologies are plagued by several issues, including a lack of realism and diversity in simulated agents, reliance on imperfect rule-based experts, and the imperative need to bridge the simulation-to-reality gap for applicability in real-world scenarios.

This paper focuses on another research direction, commonly referred to as the mid-to-mid approach, which employs post-perception results as input features. The primary advantage of this method is that the model can concentrate on learning to plan and be trained with real-world data, eliminating sim-to-real transfer concerns. Pioneering approaches such as ChauffeurNet~\cite{bansal2018chauffeurnet}, SafetyNet~\cite{vitelli2022safetynet}, and UrbanDriver~\cite{scheel2022urban} have demonstrated the ability to operate autonomous vehicles in real-world environments, with subsequent works building upon these foundations~\cite{cheng2022mpnp, pini2023safepathnet, huang2023dipp, huang2023gameformer, cheng2023plantf}. These approaches benefit significantly from advancements in the motion forecasting community, including the adoption of vector-based models~\cite{scheel2022urban, cheng2022mpnp} that excel in prediction tasks, replacing early planning models based on rasterized bird's-eye-view images~\cite{bansal2018chauffeurnet, zhou2021exploring}. However, many of these models overlook the inherent characteristics of planning tasks, such as the need for closed-loop testing and active decision-making capabilities. In contrast, our proposed framework is specifically designed for planning from the outset. Our network jointly models longitudinal and lateral driving behaviors through a query-based architecture, enabling a flexible and diverse driving style. 

Prior studies~\cite{bansal2018chauffeurnet,zhou2021exploring,cultrera2023addressing,de2019causal} have demonstrated the limitations of basic imitation learning and suggested methods for enhancement. In order to address compounding errors, an early solution can be traced back to DAgger~\cite{ross2011reduction}, which interactively refines the trained model by incorporating additional expert demonstrations. Subsequently, ChauffeurNet~\cite{bansal2018chauffeurnet} introduced perturbation-based augmentation, enabling the model to recover from minor deviations and establishing a standard practice for later research. Adding auxiliary losses such as collision loss and off-road loss~\cite{bansal2018chauffeurnet,zhou2021exploring,scheel2022urban} is another important aspect to improve the overall performance. 
However, their method is either designed for heatmap output~\cite{bansal2018chauffeurnet} or requires a differentiable rasterizer~\cite{zhou2021exploring} that converts the trajectory into a sequence of images with kernel functions. These approaches are not efficiently applicable to vector-based methods. Our research contributes to this field by introducing a novel technique that employs differentiable interpolation to bridge this gap. 

\subsection{Contrastive Learning}
Contrastive learning~\cite{hadsell2006dimensionality} is a framework that learns representation by comparing similar and dissimilar pairs, achieving significant success in computer vision~\cite{chen2020simple,he2020momentum} and natural language processing~\cite{radford2021learning}. Within the context of autonomous driving, a few attempts have been made for motion prediction. Social NCE~\cite{liu2021social} introduced a social contrastive loss to guide goal generation in pedestrian motion forecasting. Marah \textit{et al}.~\cite{halawa2022action} utilized action-based contrastive learning loss to refine learned trajectory embeddings. FEND~\cite{wang2023fend} employed this approach to recognize long-tail trajectories. These studies underscore the potency of contrastive learning in incorporating domain-specific knowledge into models through the careful selection of positive and negative examples. In our research, we extend its application to the planning domain, aiming to improve driving behavior predictions and facilitate the understanding of implicit interactions among vehicles. As generating negative samples is crucial to contrastive methods, we also introduce a new set of data augmentation functions that defines the contrastive task.
\section{Methodology}

\begin{figure*}              
\begin{center}
\includegraphics[width=0.95\linewidth]{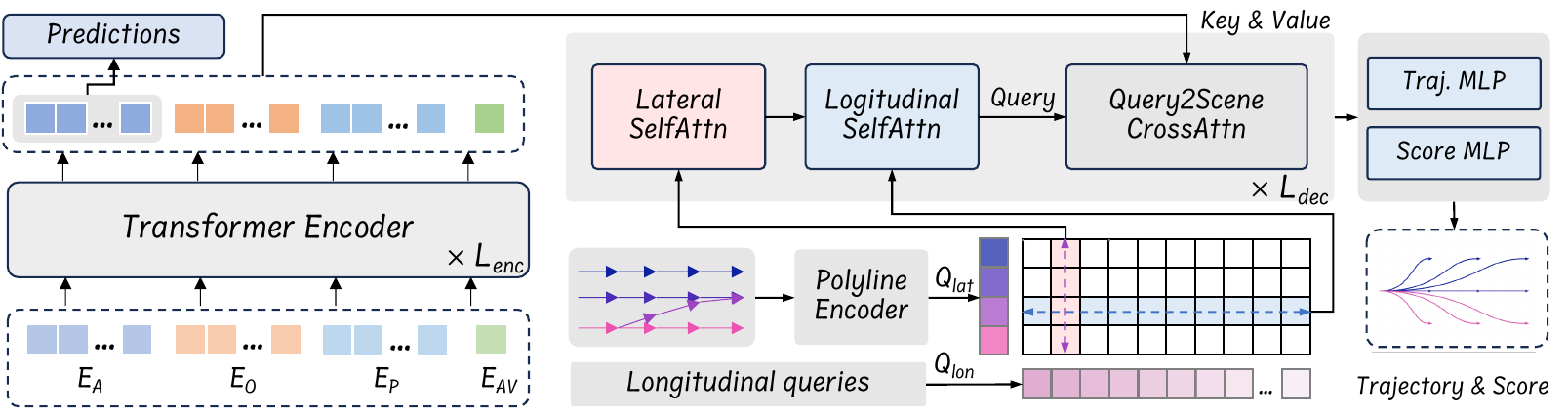}
\end{center}
\caption{
The architectural overview of the \pluto{} model is presented in this section. The model initiates lateral queries  $Q_{lat}$ using a polyline encoder based on adjacent reference lines. Simultaneously, longitudinal queries  $Q_{lon}$ are established as learnable embeddings. These queries undergo a fusion process via factorized lateral-longitudinal self-attention layers. This integration serves as a basis for the subsequent decoding of trajectories and their associated scores.
}
\label{fig:model}
\vspace{-0.5em}
\end{figure*}

\subsection{Problem Formulation}

In this study, we explore the task of autonomous driving within dynamic urban settings, considering the autonomous vehicle (AV), $N_A$ dynamic agents, $N_S$ static obstacles, a high-definition map $M$, and other traffic-related contexts $C$ such as traffic light status. We define the features of agents as $\mathcal{A} = A_{0:N_A}$, where $A_0$ represents the AV, and static obstacles are denoted by $\mathcal{O} = O_{1:N_S}$. Additionally, we denote the future state of agent $a$ at time $t$ as $\bm{y}_a^t$, with the historical and future horizons represented by $T_H$ and $T_F$, respectively. Our proposed system, \pluto{}, is designed to simultaneously generate $N_T$ multi-modal planning trajectories for the AV and a prediction for each dynamic agent. The selection of the final output trajectory, $\tau^*$, is executed by a scoring module, $\mathcal{S}$, which integrates learning-based outcomes with all scene contexts. \pluto{} is formulated as follows:
\begin{equation}
\begin{aligned}
    &(\bm{T}_{0}, \bm{\pi}_{0}), \bm{P}_{1:N_A} = f(\mathcal{A}, \mathcal{O}, M, C\,|\,\phi),\\
    &(\tau^*, \pi^*) = \underset{(\tau,\pi)\in (\bm{T}_0,\bm{\pi}_0)}{\text{arg\,max}}\, \mathcal{S}(\tau, \pi, \bm{P}_{1:N_A}, \mathcal{O}, M, C),
\end{aligned}
\end{equation}
where $f$ denotes the neural network of \pluto{}, $\phi$ is the model parameters, $(\bm{T}_{0},\bm{\pi}_{0}) = \{(\bm{y}_{0,i}^{1:T_F}, \pi_i)\,|\,i=1\dots N_T\}$ is AV's planning trajectories and corresponding confidence scores, $\bm{P}_{1:N_A} = \{\bm{y}_a^{1:T_F}\,|\,a={1\dots N_A}\}$ are agents' predictions. The subsequent sections provide a detailed illustration of each component within the \pluto{} framework.

\subsection{Input Representation and Scene Encoding}
\label{sec:encoder}

\paragraph{Agent History Encoding.} 
The observational state of each agent at any given time $t$ is denoted as $\bm{s}_i^t=\left(\bm{p}^t_i, \theta^t_i, \bm{v}^t_i, \bm{b}_i^t, \mathbb{I}^t_i\right)$, where $\bm{p}$ and $\theta$ represent the agent's position coordinates and heading angle, respectively; $\bm{v}$ refers to the velocity vector, $\bm{b}$ defines the dimensions (length and width) of the perception bounding box; and $\mathbb{I}$ is a binary indicator signifies the observation status of this frame. We convert the history sequence into vector form by calculating the difference between consecutive time steps: $\bm{\hat{s}}_i^t=\left(\bm{p}_i^t-\bm{p}_i^{t-1}, \theta^t_i - \theta^{t-1}_i ,\bm{v}_i^t-\bm{v}^{t-1}_i, \bm{b}_i^t, \mathbb{I}^t_i\right)$, resulting in agent's feature vector $F_A\in\mathbb{R}^{N_A\times{(T_H-1)}\times8}$. To extract and condense these historical features, we employ a neighbor attention-based Feature Pyramid Network (FPN)~\cite{cheng2023forecast}, which produces an agent embedding $E_A\in\mathbb{R}^{N_A\times D}$, with $D$ representing the dimensionality of the hidden layers used consistently throughout this paper.

\paragraph{Static Obstacles Encoding.} 
In contrast to motion forecasting tasks where static obstacles are often overlooked, the presence of static obstacles is crucial for ensuring safe navigation. Static obstacles encompass any entities that an AV must not traverse, such as traffic cones or barriers. 
Each static obstacle within the drivable area is represented by $\bm{o}_i=\left(\bm{p}_i, \theta_i, \bm{b}_i\right)$. We use a two-layer multi-layer-perceptron (MLP) to encode static objects features $\bm{F}_O\in\mathbb{R}^{N_S\times5}$, resulting in embedding $\bm{E}_O\in\mathbb{R}^{N_S\times D}$.

\paragraph{AV's State Encoding.} 
Drawing on insights from previous studies~\cite{cheng2023plantf,wen2020fighting} that imitation learning tends to adopt shortcuts from historical states, thereby detrimentally affecting performance, our approach only utilize the current state of AV as the input feature. This current state encompasses the AV's position, heading angle, velocity, acceleration, and steering angle. To encode the state feature while avoiding the generation of trajectories based on extrapolated kinematic states, we employ an attention-based state dropout encoder (SDE), as suggested in~\cite{cheng2023plantf}. The encoded AV's embedding is $\bm{E}_{AV}\in\mathbb{R}^{1\times D}$.

\paragraph{Vectorized Map Encoding.}
\label{sec:method_map_encoding}
The map consists of $N_P$ polylines. These polylines undergo an initial subsampling process to standardize the quantity of points, followed by the computation of a feature vector for each point. Specifically, for each polyline, the feature of the $i$-th point encompasses eight channels: $(\bm{p}_i-\bm{p}_0, \bm{p}_i - \bm{p}_{i-1}, \bm{p}_i-\bm{p}_i^{\text{left}}, \bm{p}_i-\bm{p}_i^{\text{right}})$. Here, $\bm{p}_0$ denotes the initial point of the polyline, while $\bm{p}_i^{\text{left}}$ and $\bm{p}_i^{\text{right}}$ represent the left and right boundary points of the lane, respectively. Incorporating the boundary feature is crucial as it conveys information about the drivable area, essential for planning tasks. The features of the polylines are represented as $F_P\in\mathbb{R}^{N_P\times n_p \times 8}$, where $n_p$ denotes the number of points per polyline. To encode the map features, a PointNet-like~\cite{qi2017pointnet} polyline encoder is employed, resulting in an encoded feature space $E_P\in \mathbb{R}^{N_P\times D}$.

\paragraph{Scene Encoding.}
To effectively capture the intricate interactions among various modal inputs, we concatenate different embeddings into a single tensor $E_0 \in \mathbb{R}^{(N_A+N_S+N_P+1) \times D}$. This tensor is subsequently integrated using a series of $L_{enc}$ Transformer encoders. Due to the vectorization process, the input features are stripped of their global positional information. To counteract this loss, a global positional embedding, denoted as $PE$, is introduced to each embedding. Following~\cite{zhou2023query}, $PE$ represents the Fourier embedding of the global position $(\bm{p}, \theta)$, utilizing the most recent positions of agents and static obstacles as well as the initial point of polylines. Additionally, to encapsulate inherent semantic attributes such as agent types, lane speed limits, and traffic light statuses, learnable embeddings $E_{attr}$ are incorporated alongside the input embeddings. 
$E_0$ is initialized as
\begin{align}
    E_0 &= \texttt{concat}(E_{AV}, E_A, E_O, E_P) + PE + E_{attr}.
\end{align}
The $i$-th layer of the Transformer encoder is formulated as 
\begin{equation}
\begin{aligned}
    &\hat{E}_{i-1} = \texttt{LayerNorm}({E_{i-1}}),\\
    &E_i = E_{i-1} + \texttt{MHA}(\hat{E}_{i-1},\hat{E}_{i-1},\hat{E}_{i-1}), \\
    &E_i = E_{i} + \texttt{FFN}\left(\texttt{LayerNorm}(E_i)\right),
\end{aligned}
\label{eq:encoder_sa}
\end{equation}
where $\texttt{MHA}(q,k,v)$ is the standard multi-head attention~\cite{vaswani2017attention} function, $\texttt{FFN}$ is the feedforward network layer. 
We denote $E_{enc}$ as the output of the final layer of the enoder. 

\subsection{Multi-modal Planning Trajectory Decoding}
\label{sec: decoder}

The task of planning in autonomous driving is inherently multimodal, as there are often multiple valid behaviors that could be adopted in response to a given driving scenario. For instance, a vehicle might either continue to follow a slower vehicle ahead or opt to change lanes and overtake it. To address this complex issue, we utilize a query-based, DETR-like~\cite{carion2020end} trajectory decoder. However, directly implementing the learned anchor-free queries has been found to result in mode collapse and training instability, as evidenced in~\cite{liu2021multimodal}. Drawing inspiration from the observation that driving behaviors can be decomposed into combinations of lateral (\eg, lane changing) and longitudinal (\eg, braking and accelerating) actions, we introduce a semi-anchor-based decoding structure. An illustrative overview of our decoding pipeline is presented in Fig. \ref{fig:model}, with subsequent paragraphs detailing the individual components.

\paragraph{Reference Lines as Lateral Queries.}
Following the methodology outlined in~\cite{cheng2022mpnp}, this study employs reference lines as a high-level abstraction for lateral queries. Reference lines, typically derived from the autonomous vehicle's surrounding lanes on its route, serve as a critical component in conventional vehicle motion planning, guiding lateral driving behaviors. Initially, we identify lane segments within a radius of $R_{ref}$ from the AV's current position. Starting from each identified lane segment, a depth-first search is conducted to explore all potential topological connections, linking their respective lane centerlines. Subsequently, these connected centerlines are truncated to a uniform length and are resampled to maintain a consistent number of points. The approach for representing and encoding the features of reference lines mirrors that of vectorized map encoding, as outlined in Section \ref{sec:method_map_encoding}. Ultimately, the embedded reference lines are utilized as the lateral query $Q_{lat} \in \mathbb{R}^{N_{R} \times D}$, where $N_{R}$ represents the number of reference lines. 

\paragraph{Factorized Lateral-longitudinal Self-Attention.}
In addition to $Q_{lat}$, we employ $N_{L}$ anchor-free, learnable queries $Q_{lon} \in \mathbb{R}^{N_L \times D}$ to encapsulate the multi-modal nature of longitudinal behaviors. Following this, $Q_{lat}$ and $Q_{lon}$ are combined to create the initial set of lateral-longitudinal queries, denoted as $Q_0 \in \mathbb{R}^{N_{R}\times N_{L}\times D}$:
\begin{equation}
Q_0 = \texttt{Projection}(\texttt{concat}(Q_{lat}, Q_{lon})),
\end{equation}
where $\texttt{Projection}$ refers to either a simple linear layer or a multilayer perceptron.
Since each query within $Q_0$ captures only the local region information pertaining to an individual reference line, we utilize self-attention mechanisms on $Q_0$ to integrate global lateral-longitudinal information across various reference lines. Nonetheless, applying self-attention directly to $Q_0$ results in computational complexity of $\mathcal{O}\left(N_{R}^2 N_L^2\right)$, which becomes prohibitively high as $N_{R}$ and $N_{L}$ increase. Drawing inspiration from similar approaches in the literature~\cite{ngiam2021scene}, we adopt a factorized attention strategy across each axis of $Q$, effectively reducing the computational complexity to $\mathcal{O}\left(N_R^2N_L + N_R N_L^2\right)$.

\paragraph{Trajectory Decoding.} 
The trajectory decoder consists of a sequence of $L_{dec}$ decoding layers, each comprising three types of attention mechanisms: lateral self-attention, longitudinal self-attention, and query-to-scene cross-attention. These processes are mathematically represented as follows:
\begin{equation}
\begin{aligned}
&Q'_{i-1} = \texttt{SelfAttn}(Q_{i-1}, \texttt{dim}=0),\\
&\hat{Q}_{i-1} = \texttt{SelfAttn}(Q'_{i-1}, \texttt{dim}=1),\\
&Q_i = \texttt{CrossAttn}(\hat{Q}_{i-1}, E_{enc}, E_{enc}).
\end{aligned}
\end{equation}
Here, $\texttt{SelfAttn}(X,\texttt{dim}=i)$ indicates the application of self-attention across the $i$-th dimension of $X$, and $\texttt{CrossAttn}(Q, K, V)$ incorporates layer normalization, multi-head attention, and a feed-forward network, analogous to the structure defined in Eq.~\ref{eq:encoder_sa}. The decoder's final output, $Q_{dec}$, is then employed to determine the AV's future trajectory points and their respective scores using two MLPs:
\begin{equation}
\bm{T}_0 = \texttt{MLP}(Q_{dec}),\; \bm{\pi}_0 = \texttt{MLP}(Q_{dec}).
\end{equation}
Each decoded trajectory point has six channels: $[p_x, p_y, \cos{\theta}, \sin{\theta}, v_x, v_y]$. Furthermore, to accommodate scenarios lacking reference lines, an additional MLP head is introduced to directly decode a single trajectory from the encoded features of the AV: 
\begin{equation}
\tau^{free} = \texttt{MLP}(E'_{AV}).
\end{equation}

\paragraph{Imitation Loss.} 
To avoid mode collapse, we employ the teacher-forcing~\cite{williams1989learning} technique during the training process. Firstly, the endpoint of the ground truth trajectory $\tau^{gt}$ is projected relative to reference lines, with the selection of the reference line closest in lateral distance serving as the target reference line. This target reference line is subsequently divided into $N_L-1$ equal segments by distance. Each segment corresponds to the region managed by each longitudinal query, with the final query accounting for regions extending beyond the target reference line. The query encompassing the projected endpoint is designated as the target query. By integrating the target reference line with the target longitudinal query, we derive the target supervision trajectory, $\hat{\tau}$. For trajectory regression, we employ the smooth L1 loss~\cite{girshick2015fast}, and for score classification, we utilize the cross-entropy loss, expressed as follows:
\begin{equation}
\begin{aligned}
\mathcal{L}_{reg} &= \texttt{L1}_{smooth}(\hat{\tau}, \tau^{gt}) + \texttt{L1}_{smooth}(\tau^{free}, \tau^{gt}),\\
\mathcal{L}_{cls} &= \texttt{CrossEntropy}(\bm{\pi}_{0}, \bm{\pi}^*_{0}),
\end{aligned}
\end{equation}
where $\bm{\pi}_{0}^*$ signifies the one-hot distribution derived from the index of $\hat{\tau}$. The overall imitation loss is formulated as the sum of these two components, each weighted equally:
\begin{equation}
\mathcal{L}_i = \mathcal{L}_{reg} + \mathcal{L}_{cls}.
\end{equation}

\paragraph{Prediction Loss.}
A simple two-layer MLP is used to generate a single modal prediction for each dynamic agent from the encoded agents' embeddings:
\begin{equation}
    \bm{P}_{1:N_A} = \texttt{MLP}(E'_A).
\end{equation}
Firstly, this provides dense supervision which benefits the training~\cite{shi2022motion, cheng2023forecast}. 
Secondly, the generated predictions play a crucial role in eliminating unsuitable planning proposals during the post-processing stage, as detailed in Section \ref{sec:post}.  
Denote agent's ground truth trajectory as $\bm{P}_{1:N_A}^{gt}$, the prediction loss is 
\begin{equation}
    \mathcal{L}_{p} = \texttt{L1}_{smooth}(\bm{P}_{1:N_A}, \bm{P}_{1:N_A}^{gt}).
\end{equation}

\begin{figure*}              
\begin{center}
\includegraphics[width=1.0\linewidth]{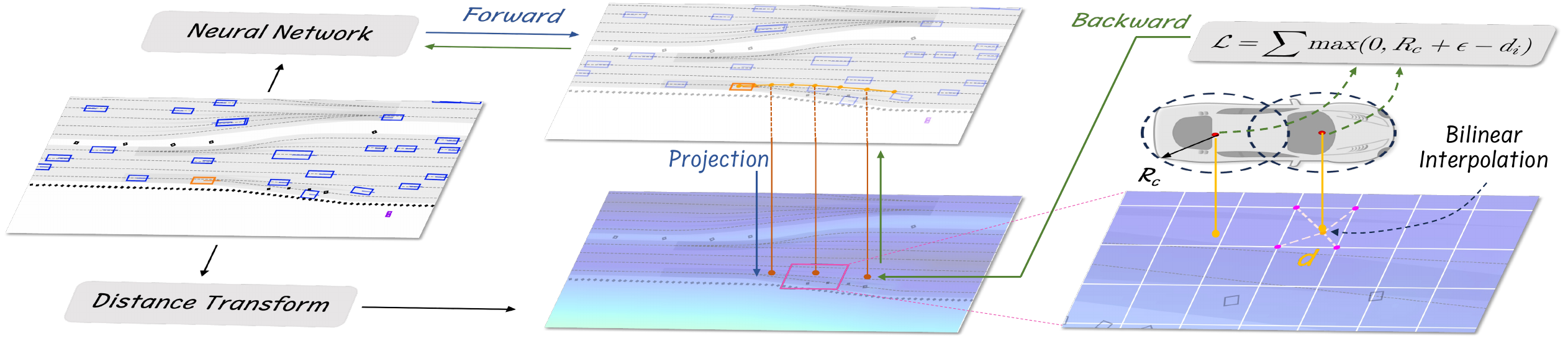}
\end{center}
\caption{
Illustration of the proposed auxiliary loss computation method. Initially, the trajectory produced by the neural network is mapped onto the image space associated with the cost map. Subsequently, the cost value is obtained via bilinear interpolation and employed in the formation of the loss function. Given the differentiable nature of all processes involved, it is feasible to incorporate auxiliary tasks directly into the framework, allowing for end-to-end training.
}
\label{fig:aux_loss}
\end{figure*}

\subsection{Efficient Differentiable Auxiliary Loss}
\label{Sect:aux}
%

As highlighted by earlier studies~\cite{bansal2018chauffeurnet,lu2022imitation_not_enough}, pure imitation learning does not suffice to preclude undesired outcomes, such as collisions with stationary obstacles or deviations from the drivable path. Therefore, it is essential to incorporate these constraints as auxiliary losses in the model during its training phase. Nevertheless, the integration of these constraints in a manner that is differentiable and enables end-to-end training of the model presents a significant challenge. A frequently adopted method for this purpose is differentiable rasterization. For instance, 
Zhou \textit{et al.} \cite{zhou2021exploring} demonstrates a technique where each trajectory point is converted into rasterized images, using a differentiable kernel function, and subsequently calculates the loss using obstacle masks within the image space. This method, however, is limited by its computational and memory demands, which in turn limits the output resolution (\eg, it permits only large time intervals and short planning horizons). To mitigate these limitations, we propose a novel approach based on differentiable interpolation. This method facilitates the concurrent calculation of auxiliary loss for all trajectory points. We take the drivable area constraint an example to elucidate our proposed method.

\paragraph{Cost Map Construction.} 
The first step of our methodology involves transforming the constraint into a queryable cost-map representation. Specifically, for the drivable area constraint, we employ the widely recognized Euclidean Signed Distance Field (ESDF) for cost representation. This process encompasses mapping the non-drivable areas (\eg, off-road regions) onto an $H\times W$ rasterized binary mask, followed by executing distance transforms on this mask. A distinctive advantage of our approach over existing methods is its elimination of the need to render the trajectory into a series of images, thereby significantly reducing computational demands. 

\paragraph{Loss Calculation}
In accordance with established methodologies in optimization-based vehicle motion planning~\cite{cheng2022gpir}, we model the vehicle's shape using $N_c$ covering circles. The trajectory point determines the centers of these circles, which can be derived in a differentiable manner. 
As illustrated in Fig. \ref{fig:aux_loss}, for each covering circle $i$ associated with a trajectory point, we obtain its signed distance value $d_i$ through projection and bilinear interpolation. To ensure adherence to the drivable area constraint, we apply a penalty to the model when $d_i$ falls below the circle's radius $R_c$. The auxiliary loss is:
\begin{equation}
    \mathcal{L}_{aux} = \frac{1}{T_f} \sum_{t=1}^{T_f}\sum_{i=1}^{N_c}\max(0, R_c+\epsilon-d_i^t),
    \label{l_aux}
\end{equation}
where $\epsilon$ is a safety threshold. Eq. \ref{l_aux} is also applicable to punish collisions with a slight change to the cost map construction. 

In practice, $d_i$ and $\mathcal{L}_{aux}$ can be differentiably and efficiently calculated batch-wise with the modern deep learning framework as shown in Algorithm \ref{alg:da_loss}. 
It is important to note that our approach is versatile and not confined to ESDF-based representations alone. Any cost representation that allows for continuous querying, such as potential fields, can be incorporated with an appropriately designed loss function.

\begin{figure}[t]
\vspace{-5pt}
\centering
\begin{algorithm}[H]
\caption{{Drivable Area Loss Pseudo-code}}
\label{alg:da_loss}
\definecolor{codeblue}{rgb}{0.25,0.5,0.25}
\definecolor{light-gray}{gray}{0.80}
\lstset{
    backgroundcolor=\color{white},
    basicstyle=\fontsize{9pt}{9pt}\ttfamily\selectfont,
    columns=fullflexible,
    breaklines=true,
    commentstyle=\fontsize{9pt}{9pt}\color{codeblue},
    keywordstyle=\fontsize{9pt}{9pt},
}
\begin{lstlisting}[language=python,mathescape,escapechar=@]
# Pytorch style pseudo-code
# traj: Trajectory, [B, T, 4] (x, y, cos, sin)
# offset: constant offset of the centers
# sdf: Signed Distance Feild [B, H, W, 1]
# res: rasterization resolution of the SDF
# Rc: radius of the covering circle
# epsilon: safety threshold 

def DriableAreaLoss(traj, sdf, offset, res, Rc, epsilon):
    H, W = sdf.shape[1:3]
    centers = traj[..., None, :2] + offset * traj[..., None, 2:4]

    # projection
    centers_pixel = torch.stack([centers[..., 0] / resolution, -centers[..., 1] / resolution],dim=-1) 
    grid = centers_pixel / torch.tensor([W//2, H//2])

    # @\textcolor{blue}{query distance in batch}@
    distance = @\textcolor{blue}{F.sample\_grid}@ (sdf.unsqueeze(1), grid, mode="bilinear").squeeze(1)
    
    # Hinge loss
    cost = Rc + epsilon - distance
    loss_mask = cost > 0
    cost.masked_fill_($\sim$loss_mask, 0)
    loss = F.l1_loss(cost, torch.zeros_like(cost), reduction="none").sum(-1)
    loss = loss.sum() / (loss_mask.sum() + 1e-6)

    return loss
\end{lstlisting}
\end{algorithm}
\end{figure}

\subsection{Contrastive Imitation Learning Framework}

\begin{figure*}              
\begin{center}
\includegraphics[width=1.0\linewidth]{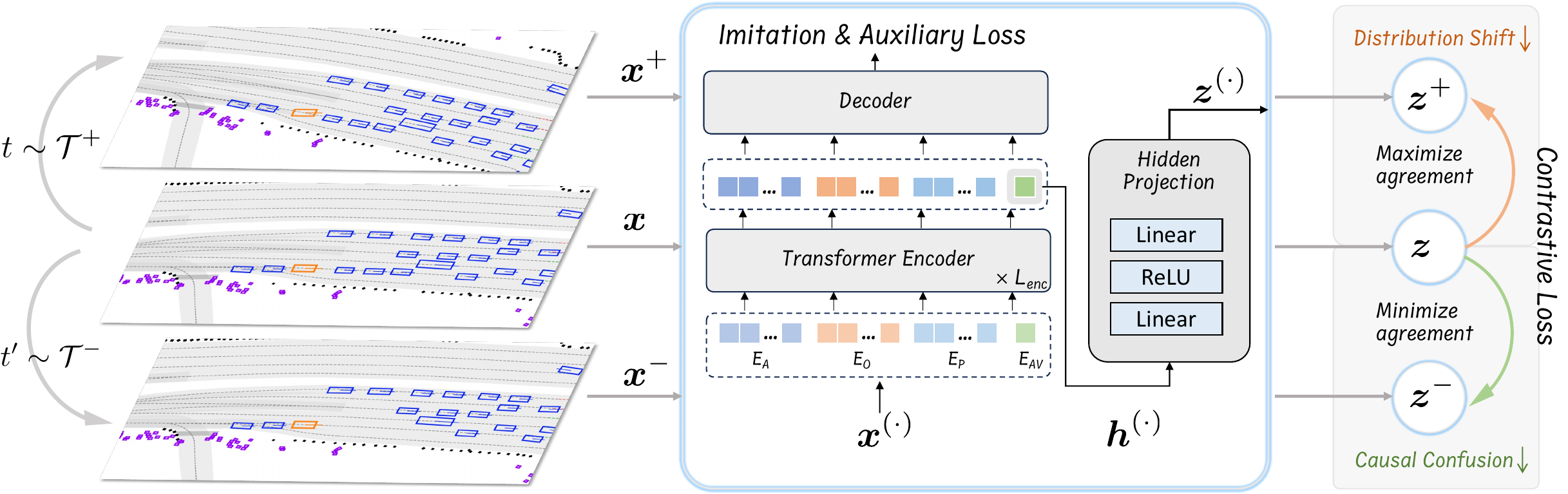}
\end{center}
\caption{
Illustration of the proposed contrastive imitation learning (CIL) framework. For any input data, we apply two data augmentation functions from different augmentation modules ($t\sim\mathcal{T}^+$ and $t'\sim\mathcal{T}^-$) to obtain a positive sample and a negative sample. The projected latent embeddings $\bm{z}^{(\cdot)}$ are used to calculate the conservative loss which maximizes the agreement between $\bm{z}^+$ and $\bm{z}$ and minimizes the agreement of $\bm{z}^-$ and $\bm{z}$.
}
\label{fig:cil}
\end{figure*}



\begin{figure*}              
\begin{center}
\includegraphics[width=1.0\linewidth]{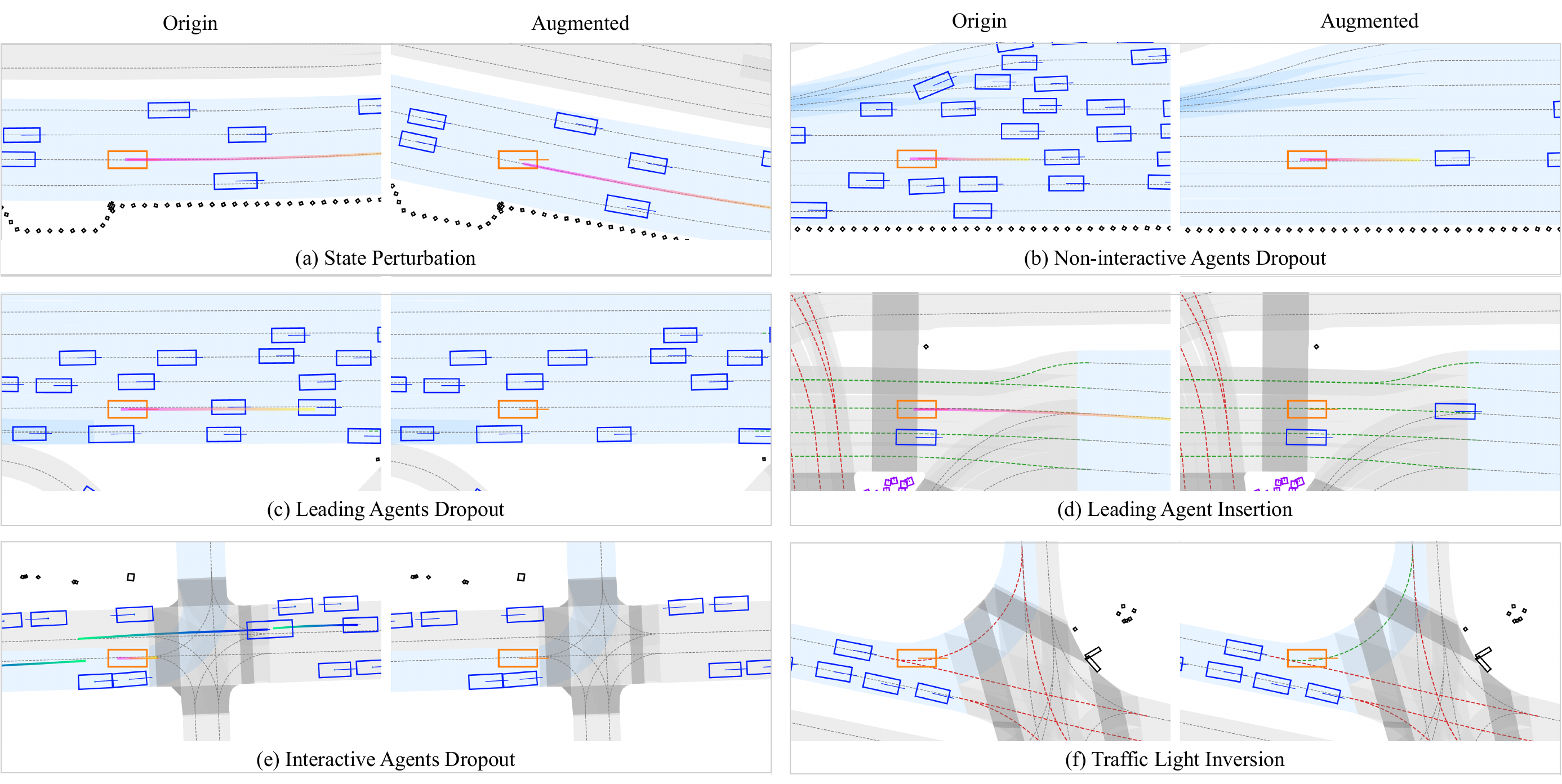}
\end{center}
\caption{Exemplary scenarios of the proposed data augmentations. 
In each group, the figure on the left denotes the origin scenario and the right one shows the augmented scene. 
AV is marked in the \ulcolor[orange]{orange}, other vehicle agents are in \ulcolor[blue]{blue}, and dash-colored lines on the lanes denote the traffic light status. (a)-(b) belongs to the positive augmentations $\mathcal{T}^+$ and (c)-(f) are negative augmentations $\mathcal{T}^-$. }
\label{fig:augs}
\end{figure*}

We introduce the Contrastive Imitation Learning (CIL) framework, designed to effectively address the challenges of distribution shift and causal confusion within a coherent and straightforward structure. Illustrated in Fig. \ref{fig:cil}, the CIL framework comprises four essential steps:

\begin{itemize}[leftmargin=*]
\item Given a training scenario data sample, denoted as $\bm{x}$, we apply both a positive data augmentation module, $\mathcal{T}^+$, and a negative one, $\mathcal{T}^-$, to generate a positive sample $\bm{x}^+$ and a negative sample $\bm{x}^-$. Positive augmentations are those that preserve the validity of the original ground truth (\eg, see Fig. \ref{fig:augs}a), while negative augmentations alter the original causal structure, rendering the original ground truth inapplicable.
\item The Transformer encoder, as detailed in Sect. \ref{sec:encoder}, is utilized to derive the latent representations $\bm{h}^{(\cdot)}$ of both the original and augmented data samples. Subsequently, these representations are mapped to a new space, represented as $\bm{z}, \bm{z}^+, \bm{z}^-$, by a two-layer MLP projection head.
\item A triplet contrastive loss is calculated to enhance the agreement between $\bm{z}$ and $\bm{z}^+$ while decreasing the similarity between $\bm{z}$ and $\bm{z}^-$.
\item Finally, trajectories for the original and positively augmented data samples are decoded, and both the imitation loss and an auxiliary loss are computed.
\end{itemize}

In practice, we randomly sample a minibatch of $N_{bs}$ samples. 
Each sample undergoes positive and negative augmentation, executed by augmentors randomly chosen from the sets $\mathcal{T}^+$ and $\mathcal{T}^-$, respectively. This augmentation triples the total number of samples to $3N_{bs}$. 
All samples are processed by the same encoder and projection head. 
Let $\text{sim}(\bm{u},\bm{v})=\bm{u}^T\bm{v}/\norm{\bm{u}}\norm{\bm{v}}$ denote the dot product between the $l2$ normalized $\bm{u}$ and $\bm{v}$, the softmax-based triple contrastive loss~\cite{goldberger2004neighbourhood} is defined as: 
\begin{equation}
    \mathcal{L}_{c} = -\log\frac{\exp\left({\text{sim}(\bm{z},\bm{z}^+)/\sigma}\right)}{\exp{\left(\text{sim}(\bm{z}, \bm{z}^+)/\sigma\right)} + \exp{\left(\text{sim}(\bm{z}, \bm{z}^-)/\sigma\right)}},
\end{equation}
where $\sigma$ denotes the temperature parameter. 
The contrastive loss is computed across all triplets in the mini-batch. 
Besides this, we provide supervision to both the original and positively augmented samples using the unmodified ground truth trajectory. 
Note that negatively augmented samples are only used to calculate the contrastive loss as their origin ground truth may be invalid after augmentation. 
The overall training loss comprises four components: imitation loss, prediction loss, auxiliary loss, and contrastive loss, represented as:
\begin{equation}
    \mathcal{L} = w_1 \mathcal{L}_i + w_2\mathcal{L}_p + w_3\mathcal{L}_{aux} + w_4\mathcal{L}_{c}.
\end{equation}

\paragraph{Data augmentations.} 
Data augmentation is the key for contrastive learning to work. 
While perturbation-based augmentations are prevalent, alternative augmentation strategies remain insufficiently explored. In this context, we present six carefully crafted augmentation functions that defines the contrastive task, with illustrative examples provided in Fig. \ref{fig:augs}.
\begin{enumerate}[leftmargin=*,label={(\arabic*)}]
\item \textit{State Perturbation} $\in \mathcal{T}^+$ (Fig. \ref{fig:augs}a): introduces minor, randomly generated disturbances to the autonomous vehicle's current position, velocity, acceleration, and steering angle. This augmentation is intended to enable the model to learn recovery strategies for slight deviations from the training distribution. The CIL framework aims to maximize the similarity between the latent representations of original and augmented samples, thereby enhancing the model's resilience to error accumulation.
\item \textit{Non-interactive Agents Dropout} $\in \mathcal{T}^+$ (Fig. \ref{fig:augs}b): omits agents from the input scenario that do not interact with the AV in the near future. Interactive agents are identified through the intersection of their future bounding boxes with the AV's trajectory. This augmentation prevents the model from learning behaviors by mimicking non-interactive agents, thereby encouraging the model to discern genuine causal relationships with interactive agents.
\item \textit{Leading Agents Dropout} $\in \mathcal{T}^-$ (Fig. \ref{fig:augs}c): removes all agents located ahead of the AV. Special consideration is given to leading-following dynamics, a prevalent situation in real-world driving. This augmentation trains the model on leading-following behaviors to prevent rear-end collisions.
\item \textit{Leading Agent Insertions} $\in \mathcal{T}^-$ (Fig. \ref{fig:augs}d): introduces a leading vehicle into the AV's original path, at a position where the AV's expected trajectory becomes invalid (\eg, would result in a collision). The inserted vehicle's trajectory data is sourced from a randomly selected agent in the current mini-batch to maintain data realism.
\item \textit{Interactive Agent Dropout} $\in \mathcal{T}^-$ (Fig. \ref{fig:augs}e): excludes agents that have direct or indirect interactions with the AV. Identification of interactive agents follows the methodology outlined in \textit{Non-interactive Agents Dropout}. This function aims to train the model on less intuitive interactions within complex scenarios, such as unprotected left turns and lane changes.
\item \textit{Traffic Light Inversion} $\in \mathcal{T}^-$ (Fig. \ref{fig:augs}f): in scenarios where the AV approaches an intersection governed by traffic lights without a leading vehicle, the traffic light status is reversed (\eg, from red to green). This function teaches the model to adhere to basic traffic light rules.
\end{enumerate}
These augmentation functions are designed with minimal inductive bias to ensure broad applicability. The contrastive learning task facilitates an implicit feedback mechanism, providing implicit reward signals. These signals reinforce adherence to fundamental driving principles. 

\subsection{Planning and Post-processing}
\label{sec:post}

\begin{algorithm}[t]
    \caption{Trajectory planning process of \pluto{}}
    \label{alg:planning}
    \label{euclid}
    \begin{algorithmic}
        \State \textbf{Input:} Init state $\bm{y}_0$, scenario feature $\bm{x}$, constant $K, \alpha, N_T$
        \Procedure{Trajectory Selection}{}
        \State $\bm{T}_0, \bm{\pi}_{0}, \bm{P}_{1:N_A} = \texttt{\pluto{}}(\bm{x})$ \Comment{\textcolor{gray}{Run model inference}}\vskip 3pt
        \State $\bm{T}_0 = \texttt{TopK}(\bm{T}_0, \bm{\pi}_{0}, K)$
        \Comment{\textcolor{gray}{Select Top-K trajectories}}\vskip 3pt
        \State Initialize rollouts $\widetilde{T}_0 = [\bm{y}_0]$
        \For{$t$ in $1\dots N_T$} \Comment{\textcolor{gray}{Forward Simulation}}
            \State $\bm{u}_0 = \texttt{LQRTracker}(\bm{y}_0, \bm{T}_0, t)$
            \State $\bm{y}_0 = \texttt{BicycleModel}(\bm{y}_0, \bm{u}_0)$\vskip 3pt
            \State $\widetilde{\bm{T}}_0 = [\widetilde{\bm{T}}_0, \bm{y}_0]$
        \EndFor
        \State $\bm{\pi}_{rule}$ = \texttt{RuleBasedEvaluator}($\widetilde{T}_0, \bm{P}_{1:N_A}, \bm{x}$)
        \State $\tau^* = \bm{T}_0[i], \; i = \text{argmax}(\bm{\pi}_{rule} + \alpha\bm{\pi}_{0})$
         \State \Return $\tau^*$
        \EndProcedure
    \end{algorithmic}
\end{algorithm}

In the context of trajectory planning, our objective is to select a deterministic future trajectory from the diverse outcomes provided by the multi-modal outputs, as discussed in Section \ref{sec: decoder}. Rather than merely selecting the most likely trajectory, we integrate a post-processing module to serve as an additional safety verification mechanism, as illustrated in Algorithm \ref{alg:planning}.

Upon extracting the scenario's features, the model is executed to generate multi-modal planning trajectories $\bm{T}_0 \in \mathbb{R}^{N_R N_L\times T_F\times6}$, associated confidence scores $\bm{\pi}_{0} \in \mathbb{R}^{N_R N_L}$, and predictions for the agents' movements $\bm{P}_{1:N_A} \in \mathbb{R}^{N_A \times T_F \times 2}$. Given that the total trajectory count $N_R N_L$ for $\bm{T}_0$ can be extensive, an initial filtering step retains only the top $K$ trajectories, ranked by their confidence scores, to streamline subsequent computations.

Following~\cite{Dauner2023CORL}, a closed-loop forward simulation is performed on $\bm{T}_0$ to obtain simulated rollouts $\widetilde{\bm{T}}_0$, utilizing a linear quadratic regulator (LQR) for trajectory tracking and a kinematic bicycle model for state updates. It has been noted~\cite{cheng2023plantf} that trajectory-based imitation learning may not fully account for the dynamics of the underlying system, potentially leading to discrepancies between the model's planned trajectory and its actual execution. To mitigate this issue, our assessment relies on the simulated rollouts rather than the model's direct output, thus narrowing the gap.

Subsequently, a rule-based evaluator assigns scores $\bm{\pi}_{rule}$ to each simulated rollout based on criteria such as progress, driving comfort, and adherence to traffic regulations, in alignment with the framework established in~\cite{Dauner2023CORL}. This evaluation also incorporates predictions of agents' trajectories $\bm{P}_{1:N_a}$ to calculate the time-to-collision (TTC) metric, excluding rollouts that result in at-fault collisions. The ultimate score combines the initial learning-based confidence score $\bm{\pi}_{0}$ with the rule-based score $\bm{\pi}_{rule}$ via the equation:

\begin{equation}
\bm{\pi} = \bm{\pi}_{rule} + \alpha \bm{\pi}_{0},
\label{eq:final_score}
\end{equation}
where $\alpha$ represents a fixed weighting factor. The selection of the final trajectory $\tau^*$ is based on maximizing $\bm{\pi}$. 
Unlike the post-processing step described in~\cite{huang2023gameformer, huang2023dipp}, which typically utilizes an optimizer to refine the trajectory, our post-processing module acts solely as a trajectory selector, leaving the original planning trajectory unaltered.
We regard the post-processing step as a proxy to inject human preference or control into the black-boxed neural network, acknowledging its current limitations, and providing a lower-bound safety assurance to mitigate the risk of catastrophic accidents.

\section{Experiments}

\subsection{Experiment Setup}

\paragraph{nuPlan.}
Our model was trained and evaluated using the nuPlan dataset~\cite{caesar2021nuplan}. This dataset comprises 1,300 hours of real-world driving data, encompassing up to 75 labeled scenario types. It introduces the first publicly accessible, large-scale planning benchmark for autonomous driving through its associated closed-loop simulation framework. Each simulation conducts a 15-second rollout at a frequency of 10 Hz, during which the autonomous vehicle is managed by a planner and tracker. Traffic agents within these simulations are controlled in two distinct manners: \textit{non-reactive}, wherein agents' states are determined based on logged trajectories, and \textit{reactive}, wherein agents are governed by an Intelligent Driver Model~\cite{treiber2000idm} planner.

\paragraph{Benchmark and Metrics.} 
For all experiments, we use a standardized training split of 1M frames sampled from all scenario types. 
For evaluation, we use the \textbf{Val14} benchmark~\cite{Dauner2023CORL}, which contains up to 100 scenarios from the 14 scenario types specified in the nuPlan planning challenge, resulting in a total number of 1090 scenarios (we filter out a few scenarios that initialized as failed in the reactive simulations). 

NuPlan employs three principal evaluation metrics: the open-loop score (OLS), the non-reactive closed-loop score (NR-score), and the reactive closed-loop score (R-score). Given that previous studies have demonstrated a minimal correlation between open-loop prediction performance and closed-loop planning effectiveness, we only focus on closed-loop performance in this paper. The closed-loop score is calculated as a weighted average of several key metrics:
\begin{enumerate}[leftmargin=*,label={(\arabic*)}]
    \item \textit{No ego at-fault collisions}: A collision is identified when the autonomous vehicle's (AV) bounding box intersects with that of other agents or static obstacles. However, collisions initiated by other agents, such as rear-end collisions, are disregarded.
    \item \textit{Time to collision (TTC) within bound}: The TTC is defined as the time it would take for the AV and another entity to collide if they continue on their current trajectories and speeds. This metric mandates that the TTC exceeds a specified threshold.
    \item \textit{Drivable area compliance}: This criterion requires that the AV remains within the boundaries of the drivable roadway at all times, ensuring adherence to the designated driving area.
    \item \textit{Comfortableness}: The comfort of the AV is quantified by examining the minimum and maximum values of its longitudinal and lateral accelerations and jerks, as well as its yaw rate and acceleration. These parameters are assessed against established thresholds derived from empirical data.
    \item \textit{Progress}: Progress is evaluated by comparing the distance covered by the AV along its planned route to that achieved by an expert driver, expressed as a percentage.
    \item \textit{Speed limit compliance}: This metric checks whether the AV's speed falls within the legal limits prescribed for the roadway it is traversing.
    \item \textit{Driving direction compliance}: This measure penalizes deviations from the correct driving direction, particularly incidents where the AV is found traveling against the flow of traffic.
\end{enumerate}
A more detailed description and calculation of the metrics can be found at~\cite{nuplan2023metrics}. 
We use the non-reactive closed-loop score (denoted as \textbf{score} if not specified) as our primary overall performance evaluation metric. 

\paragraph{Baselines.}
In this study, we conduct a comparative analysis between \pluto{} and both existing and state-of-the-art (SOTA) methodologies utilizing the nuPlan benchmark to demonstrate the efficacy of our proposed method. The baselines for comparison are categorized into three groups: rule-based, pure learning, and hybrid approaches. Rule-based methods rely on manually engineered rules without incorporating learning processes. In contrast, pure learning methods employ neural networks to directly generate the final planned trajectory, omitting any refinement or post-processing stages. Hybrid methods, however, include a post-processing module to refine or adjust outcomes derived from learning-based techniques.
The benchmarked methods are outlined as follows:

\begin{enumerate}[leftmargin=*,label={(\arabic*)}]
\item \textbf{Intelligent Driver Model} (\textbf{IDM})~\cite{treiber2000idm}: This is a classic, time-continuous car-following model extensively utilized in traffic simulations. We employ the official implementation as referenced in the literature \cite{caesar2021nuplan}.
\item \textbf{PDM-Closed}~\cite{Dauner2023CORL}: Identified as the winning entry in the 2023 nuPlan planning challenge, this method generates a series of proposals by integrating IDM policies with varying hyperparameters, subsequently selecting the optimal one through a rule-based scoring system. Despite its simplicity, it has proven effective in practice and currently holds the SOTA performance. Its open-source implementation is utilized in our study.
\item \textbf{PDM-Open}~\cite{Dauner2023CORL}: This approach, centered around a predictive model that conditions on the centerline and utilizes MLPs, is implemented through an available open-source version.
\item \textbf{GC-PGP}~\cite{hallgarten2023gc-pgp}: A predictive model that focuses on goal-conditioned lane graph traversals.
\item \textbf{RasterModel}: A CNN-based model that interprets the input scenario as a multi-channel image, as described in referenced literature~\cite{caesar2021nuplan}.
\item \textbf{UrbanDriver}~\cite{scheel2022urban}: A learning-based planner that leverages vectorized inputs through PointNet-based polyline encoders and Transformers. This model is assessed through its open-loop re-implementation, incorporating historical data perturbation during its training phase.
\item \textbf{PlanTF}~\cite{cheng2023plantf}: A strong pure imitation learning baseline that leverages a Transformer architecture to explore efficient design in imitation learning.  Despite its simplicity, it stands as the current SOTA among pure learning models.
\item \textbf{GameFormer}~\cite{huang2023gameformer}: Modeled on DETR-like interactive planning and prediction based on level-k games, the output from this model serves as an initial estimate, which is further refined through a nonlinear optimizer. The official open-source code is used for implementation purposes.
\item \textbf{PlanTF-H}: This method enhances PlanTF by integrating a post-processing module as described in Sect. \ref{sec:post}, thereby converting it into a hybrid approach.
\end{enumerate}

\subsection{Implementation Details}
We present two variations \pluto{}$^\dagger$ and \pluto{}, differing only in that \pluto{}$^\dagger$ omits the post-processing step. Feature extraction focuses on map elements and agents within a 120-meter radius of the autonomous vehicle. We adhere to the nuPlan challenge by setting the planning and historical data horizons at 8 seconds and 2 seconds, respectively. The model incorporates auxiliary tasks designed to penalize off-road driving and collisions.
Training was conducted using 4 RTX3090 GPUs, with a batch size of 128 over 25 epochs. We utilized the AdamW optimizer, applying a weight decay of $1e^{-4}$. The learning rate is linearly increased to $1e^{-3}$ over the first three epochs and then follows a cosine decay pattern throughout the remaining epochs. The loss weights $w_{1-4}$ are uniformly assigned a value of 1.0. The training finishes in 45 hours with CIL and 22 hours without it. Details on further parameter settings can be found in Table \ref{tab:params}.

\begin{table}[h]
\centering
\caption{Parameters used in \pluto{}$^\dagger$ and \pluto{}}
\label{tab:params}
\setlength{\tabcolsep}{9pt}
\renewcommand{\arraystretch}{1.1}
\small
\begin{tabular}{y{30}|y{90}|y{40}}
Notation & Parameters & Values \\ 
\specialrule{.1em}{0.5ex}{0.5ex}
$T_H$ & Historical timesteps & 20 \\
$T_F$ & Future timesteps & 80 \\
$D$ & Hidden dimension & 128 \\
$L_{enc}$ & Num. encoder layers & 4 \\
$L_{dec}$ & Num. decoder layers & 4 \\
$N_L$ & Num. lon. queries & 12 \\
$N_c$ & Num. covering circles & 3 \\
$[H, W]$ & Cost map size & $500\times500$\\
- & Cost map resolution & 0.2m \\
$\alpha$ & Score weight & 0.3 \\
$\sigma$ & Tempreture parameter & 0.1
\end{tabular}
\vspace{-1em}
\end{table}

\section{Results and Disscusion}

\begin{table*}[]
\begin{center}
\caption{Closed-loop planning results on the \textbf{Val14} benchmark. All metrics are \underline{higher the better}.}
\label{tab:val14}
\vspace{-1em}
\setlength{\tabcolsep}{9pt}
\renewcommand{\arraystretch}{1.2}
\small
\begin{tabular}{y{50}y{65}|x{28}x{28}x{28}x{28}x{28}x{28}x{28}|x{28}}
\toprule
Type & Planner & Score & Collisions & TTC & Drivable & Comfort & Progress & Speed & R-score \\ \midrule
\textcolor{gray}{Expert} & \multicolumn{1}{l|}{\textcolor{gray}{Log-Replay}} & \textcolor{gray}{93.68} & \textcolor{gray}{98.76} & \textcolor{gray}{94.40} & \textcolor{gray}{98.07} & \textcolor{gray}{99.27} & \textcolor{gray}{98.99} & \textcolor{gray}{96.47} & \textcolor{gray}{81.24} \\ \midrule
\multirow{2}{*}{Rule-based} & \multicolumn{1}{l|}{IDM\cite{treiber2000idm}} & 79.31 & 90.92 & 83.49 & 94.04 & 94.40 & 86.16 & 97.33 & 79.31 \\
 & \multicolumn{1}{l|}{PDM-Closed\cite{Dauner2023CORL}} & 93.08 & 98.07 & 93.30 & \textbf{99.82} & 95.52 & 92.13 & \textbf{99.83} & \textbf{93.20} \\ \midrule
\multirow{6}{*}{Pure Learning} & \multicolumn{1}{l|}{PDM-Open\cite{Dauner2023CORL}} & 50.24 & 74.54 & 69.08 & 87.89 & 99.54 & 69.86 & 97.72 & 54.86 \\
 & \multicolumn{1}{l|}{GC-PGP\cite{hallgarten2023gc-pgp}} & 61.09 & 85.87 & 80.18 & 89.72 & 90.00 & 60.32 & 99.34 & 54.91 \\
 & \multicolumn{1}{l|}{RasterModel\cite{caesar2021nuplan}} & 66.92 & 86.97 & 81.46 & 85.04 & 81.46 & 80.60 & 98.03 & 64.66 \\
 & \multicolumn{1}{l|}{UrbanDriver\cite{scheel2022urban}} & 67.72 & 85.60 & 80.28 & 90.83 & \textbf{100.0} & 80.83 & 91.58 & 64.87 \\
 & \multicolumn{1}{l|}{PlanTF\cite{cheng2023plantf}} & 85.30 & 94.13 & 90.73 & 96.79 & 93.67 & 89.83 & 97.78 & 77.07 \\
 & \multicolumn{1}{l|}{\pluto{}$^\dagger$ (w/o post.)} & 89.04 & 96.18 & 93.28 & 98.53 & 96.41 & 89.56 & 98.13 & 80.01 \\ \midrule
\multirow{3}{*}{Hybrid} & \multicolumn{1}{l|}{GameFormer\cite{huang2023gameformer}} & 82.95 & 94.32 & 86.77 & 94.87 & 93.39 & 89.04 & 98.67 & 83.88 \\
 & \multicolumn{1}{l|}{PlanTF-H\cite{cheng2023plantf}} & 89.96 & 97.06 & 93.38 & 97.79 & 91.08 & 92.90 & 98.01 & 88.08 \\
 & \pluto{} (Ours) & \textbf{93.21} & \textbf{98.30} & \textbf{94.04} & 99.72 & 91.93 & \textbf{93.65} & 98.20 & 92.06 \\ \bottomrule
\end{tabular}
\end{center}
\label{tab:sota}
\vspace{-1em}
\end{table*}

\subsection{Comparison with State of the Art}
The comparative analysis with other methods on the \textbf{Val14} benchmark is detailed in Table \ref{tab:val14}. Initially, our purely learning-oriented variant, \pluto{}$^\dagger$, surpasses all prior baselines dedicated to pure learning. Significantly, when compared to the leading model, PlanTF, \pluto{}$^\dagger$ demonstrates marked improvements across nearly all evaluated metrics, with particular enhancements observed in metrics pertinent to safety (\eg, \textit{Collisions} improved from $94.13$ to $96.18$, \textit{TTC} from $90.73$ to $93.28$, and \textit{Drivable} from $96.79$ to $98.53$). These results highlight the constraints of models based solely on imitation and underscore the effectiveness of incorporating auxiliary loss and the design of the CIL framework.

Furthermore, our hybrid model, \pluto{}, attains the highest scores across all baselines, surpassing the current state-of-the-art rule-based model, PDM-Closed, for the first time. This achievement emphasizes the promise of learning-based approaches in planning. Remarkably, the performance of our methods closely aligns with that of the log-reply expert (scoring $93.68$ vs. $93.21$), indicating a significant stride towards expert-level planning.

In addition to quantitative outcomes, our method also presents an advantage in driving behavior over PDM-Closed. Given that PDM-Closed primarily focuses on speed planning, its capability in lateral maneuvers is somewhat restricted, limiting its ability to execute lane-change actions. In contrast, \pluto{} is designed to consider both longitudinal and lateral movements, thanks to the query-based architecture of our model. This capability will be further illustrated through case studies in the section on qualitative results.

\begin{figure*}              
\begin{center}
\includegraphics[width=\textwidth]{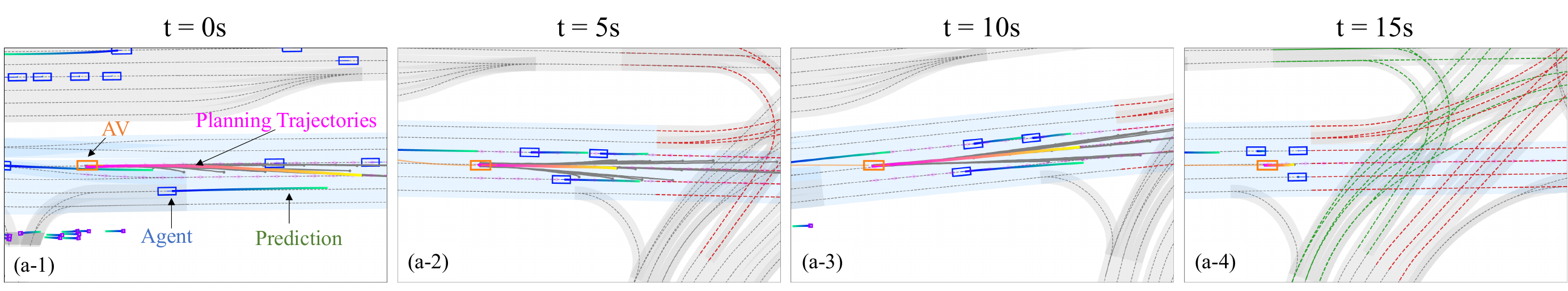} 

\vspace{1.5pt}
\includegraphics[width=\textwidth]{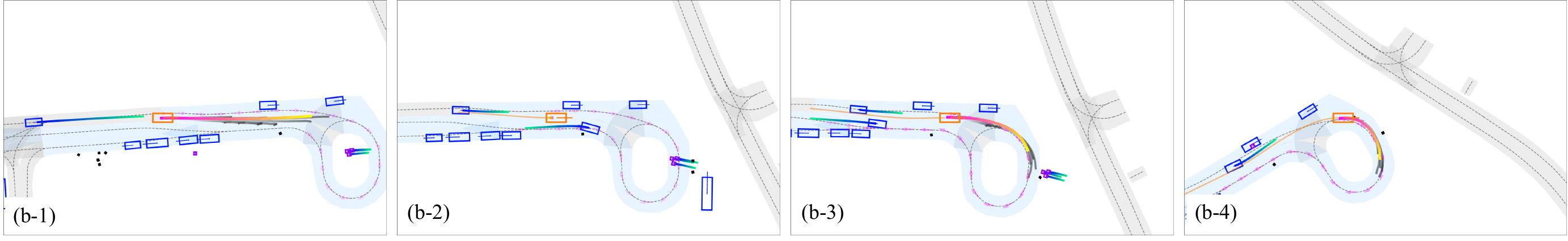}

\vspace{1.5pt}
\includegraphics[width=\textwidth]{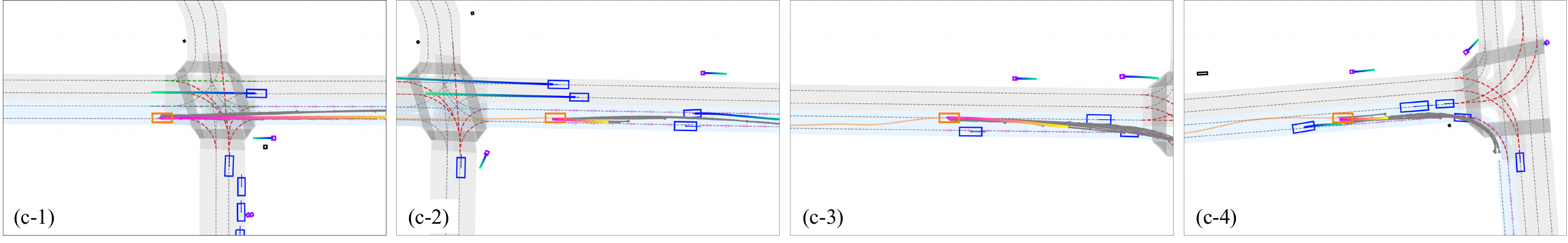}

\vspace{1.5pt}
\includegraphics[width=\textwidth]{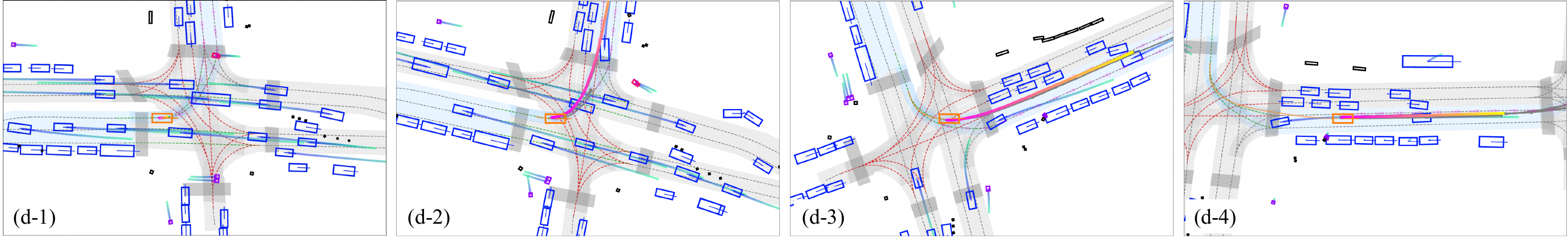}

\vspace{1.5pt}
\includegraphics[width=\textwidth]{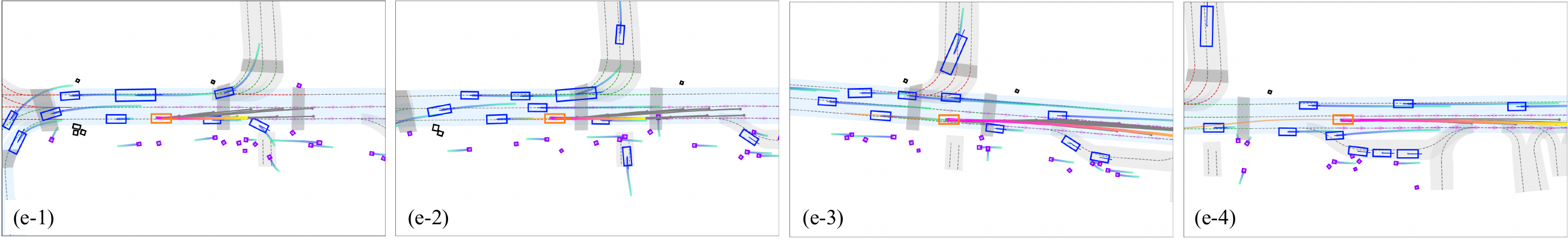}
\end{center}
\caption{Qualitative results of closed-loop planning for five representative scenarios from the test set. Each scenario (every row) lasts 15 seconds and we take 4 snapshots with a 5-second interval. Purple dash arrows denote the reference lines, the light blue lanes denote the global routing plan and other important legends are marked in Fig. a-1. It is recommended to refer to our \href{https://jchengai.github.io/pluto}{project website} for more vivid videos. }
\label{fig:qua}
\end{figure*}

\subsection{Qualitative Results}

Fig. \ref{fig:qua} presents selected scenarios from the nuPlan test set, showcasing the robust performance of our framework in complex, interactive urban driving situations through the exhibition of diverse, human-like behaviors. The scenarios are detailed as follows:

\begin{enumerate}[leftmargin=*,label={(\alph*)}]
\item The autonomous vehicle navigates to an adjacent empty lane to enhance efficiency and subsequently halts at a red light at the intersection. Observations from sequences a-2 and a-3 reveal that \pluto{} concurrently evaluates multiple potential plans for different behaviors (illustrated by gray 
candidate trajectories), enhancing the planning process's flexibility and resemblance to human driving.
\item In a scenario requiring the AV to maneuver around a roundabout, its path is narrowed by a parked vehicle. Our planning system adeptly navigates around this obstacle while yielding to an oncoming vehicle in a constrained space, exemplifying our method's competence in managing static obstacles and interacting with other vehicles.
\item Encountering a stationary vehicle within its lane, the AV executes a left lane change to bypass the obstacle, subsequently returning to its original lane to adhere to the intended route. This scenario underscores the planner's dynamic decision-making capabilities and its adeptness in route adherence and road topology comprehension.
\item During a left-turn maneuver in heavy traffic, the AV patiently waits for an opportune moment to execute the turn, illustrating our method's effectiveness in navigating intersections in high-density traffic conditions.
\item Upon following a slower vehicle, the AV opts to accelerate and overtake, showcasing a behavior that aligns closely with natural human driving and highlighting our proposed method's adaptability.
\end{enumerate}

In summary, \pluto{} exhibits advanced and varied driving behaviors unattainable through simplistic speed-planning methods (\eg, PDM-Closed). Its ability to execute natural lane changes, navigate around obstacles, and dynamically modify decisions in interactive scenarios marks a significant advancement towards the realization of practical learning-based planning. For further insights, including videos, we direct interested readers to our project website.

\subsection{Ablation Studies}
For all ablation studies, we evaluate on a subset of nuPlan (non-overlapping with the \textbf{Val14} benchmark), which contains 20 scenarios for each of the 14 scenarios types.

\paragraph{Influence of Each Component.} 
\begin{table*}[]
\begin{center}
\caption{Ablation study of the influence of each component}
\label{tab:component}
\setlength{\tabcolsep}{8pt}
\renewcommand{\arraystretch}{1.2}
\small
\begin{tabular}{y{20}y{100}|x{25}x{35}x{25}x{30}x{30}x{30}x{30}x{30}}
\toprule
Model           & Description                       & Score & Collisions & TTC   & Drivable & Comfort & Progress & Speed & Direction \\ \midrule
\;\;-- & \textcolor{gray}{PlanTF~\cite{cheng2023plantf}}  & \textcolor{gray}{87.55} & \textcolor{gray}{96.39}      & \textcolor{gray}{90.76} & \textcolor{gray}{96.79}    & \textcolor{gray}{93.57}   & \textcolor{gray}{90.50}    & \textcolor{gray}{96.75} & \textcolor{gray}{99.40}    \\
\midrule
$\mathcal{M}_0$ & Base                              & 87.04 & 95.92      & 91.43 & 95.92    & 95.92   & 90.90    & 91.01 & 100.0     \\
$\mathcal{M}_1$ & $\mathcal{M}_0$ + SDE             & 89.64 & 97.37      & 95.14 & 97.57    & 96.36   & 90.16    & 96.91 & \textbf{100.0}     \\
$\mathcal{M}_2$ & $\mathcal{M}_1$ + Auxiliary loss  & 90.03 & 97.98      & 93.52 & 98.38    & \textbf{96.76}   & 90.04    & \textbf{97.21} & 99.39     \\
$\mathcal{M}_3$ & $\mathcal{M}_2$ + Ref. free head  & 90.69 & 97.57      & 94.74 & 99.19    & 94.33   & 89.86    & 97.13 & 99.60     \\
$\mathcal{M}_4$ & $\mathcal{M}_3$ + CIL & 91.66 & 97.59 & 94.38 & 99.60 & 96.39 & 91.30 & 96.58 & 99.60 \\
$\mathcal{M}_5$ & $\mathcal{M}_4$ + Post-processing & \textbf{93.57} & \textbf{98.39}  & \textbf{95.58} & \textbf{99.60}    & 93.17   & \textbf{93.32}    & 97.08 & 99.80     \\ 
\midrule
\;\;-- & \textcolor{gray}{Expert (Log-Replay)} & \textcolor{gray}{94.24} & \textcolor{gray}{98.59} & \textcolor{gray}{95.58} & \textcolor{gray}{98.39}    & \textcolor{gray}{99.60}   & \textcolor{gray}{99.06}    & \textcolor{gray}{94.53} & \textcolor{gray}{99.80}     \\
\bottomrule
\end{tabular}
\end{center}
\label{tab:sota}
\end{table*}
Table \ref{tab:val14} shows the trajectory from a base model to the top-performing learning-based planner. The initial model, denoted as $\mathcal{M}_0$, is constructed on the architecture depicted in Sect. \ref{sec:encoder} and \ref{sec: decoder}, employing solely imitation loss for training. $\mathcal{M}_0$ demonstrates performance on par with the previous SOTA pure learning-based method, PlanTF, an achievement we ascribe to the enhanced query-based architecture.

The introduction of the state dropout encoder (SDE) in $\mathcal{M}_1$, which randomly masks the autonomous vehicle's kinematic states during training to avert the generation of shortcut trajectories by state extrapolation, results in marked improvements over $\mathcal{M}_0$ across almost all metrics.

$\mathcal{M}_2$ incorporates an auxiliary loss designed to penalize deviations from the drivable area and collisions. This modification leads to enhancements in both the \textit{Collision} metric (from 97.37 to 97.98) and the \textit{Drivable} metric (from 95.92 to 98.38). 
We would like to highlight that despite the seemingly minor difference in total scores between $\mathcal{M}_2$ and $\mathcal{M}_1$, the disparity in their actual performance is substantial. We direct interested readers to the project website for the model ablation results. 
The $\mathcal{M}_3$ model underscores the importance of integrating a reference line-free decoding head to effectively handle scenarios where reference lines are absent, such as in parking lots.

Further, $\mathcal{M}_4$ is trained using the proposed contrastive imitation learning framework, achieving a significant uplift in performance from 90.69 to 91.66. This improvement is noteworthy, particularly as it is already approaching the expert's performance.

Ultimately, $\mathcal{M}_5$ attains the best overall performance, though at a minor trade-off in the \textit{Comfort} metric. This compromise stems from the increased incidence of emergency stops triggered by the safety checker, thereby enhancing safety-related metrics.

\begin{table}[]
\begin{center}
\caption{Impact of different longitudinal queries $N_{L}$ (based on $\mathcal{M}_4$)}
\label{tab:num_lon_query}
\vspace{-0.5em}
\setlength{\tabcolsep}{7pt}
\renewcommand{\arraystretch}{1.2}
\small
\begin{tabular}{y{15}|x{28}x{28}x{28}x{28}x{28}}
\toprule
$N_L$ & Score & Collisions & TTC & Comfort & Progress \\
\midrule
6 & 88.89 & 96.56 & 93.12 & 96.36 & 89.47\\
\baseline{12} & \baseline{\textbf{91.66}} & \baseline{\textbf{97.59}} & \baseline{94.38} & \baseline{\textbf{96.39}}
& \baseline{\textbf{91.30}}\\
18 & 90.18 & 97.17 & \textbf{95.14} & 95.95 & 90.88\\
24 & 87.90 & 95.58 & 93.57 & 95.58 & 89.14\\
\bottomrule
\end{tabular}
\end{center}
\label{tab:sota}
\vspace{-1em}
\end{table}

\paragraph{Number of Longitudinal Queries}. 
Table \ref{tab:num_lon_query} presents the outcomes associated with various quantities of longitudinal queries $N_L$ (utilizing model $\mathcal{M}_4$). 
The results indicate that a setting of $N_L=12$ yields the most favorable performance among four tested variants. This suggests an appropriate number of queries is necessary to cover all the longitudinal behaviors for a planning horizon of 8s. 
An increase in $N_L$ beyond this point detracts from performance. This decline can likely be attributed to the sufficiency of $N_L=12$ in capturing a diverse array of behaviors; additional queries become redundant, potentially increasing the training difficulty. 

\begin{table}[]
\begin{center}
\caption{Impact of different $K$ in post-processing (based on $\mathcal{M}_5$)}
\label{tab:abl_topk}
\vspace{-0.5em}
\setlength{\tabcolsep}{7pt}
\renewcommand{\arraystretch}{1.2}
\small
\begin{tabular}{y{15}|x{28}x{38}x{28}x{28}x{28}}
\toprule
$K$ & Score & Collisions & TTC & Comfort & Progress \\
\midrule
10 & 93.18 & 98.19 & 95.18 & 93.57 & 92.73\\
\baseline{20} & \baseline{93.57} & \baseline{\textbf{98.39}} & \baseline{95.58} & \baseline{\textbf{93.17}} & \baseline{93.32}\\
30 & \textbf{93.58} & 98.39 & \textbf{95.98} & 91.57 & 93.66\\
40 & 93.02 & 98.39 & 94.38 & 90.76 & \textbf{93.79}\\
\bottomrule
\end{tabular}
\end{center}
\label{tab:sota}
\vspace{-1em}
\end{table}

\paragraph{Top-K in Coarse Selection}. 
In the planning cycle, \pluto{} generates a total of $N_R \times N_L$ trajectories. Empirical evidence suggests that trajectories associated with low confidence scores often exhibit inferior quality. Consequently, employing a preliminary selection process based on confidence scores proves advantageous in eliminating such trajectories, thereby expediting subsequent post-processing. As illustrated in Table \ref{tab:abl_topk}, setting $K=20$ turns out to be appropriate. 

\begin{table}[]
\begin{center}
\caption{Impact of $\alpha$ in trajectory selection (Based on $\mathcal{M}_5$)}
\label{tab:abl_alpha}
\vspace{-0.5em}
\setlength{\tabcolsep}{7pt}
\renewcommand{\arraystretch}{1.2}
\small
\begin{tabular}{y{20}|x{28}x{28}x{28}x{28}x{28}}
\toprule
$\alpha$ & Score & Collisions & TTC & Comfort & Progress \\
\midrule
Rule & 90.64 & 96.79 & 91.77 & 80.32 & 98.43\\
\midrule
0.1 & 92.27 & 97.99 & 92.71 & 85.14 & \textbf{96.48}\\
\baseline{0.3} & \baseline{\textbf{93.57}} & \baseline{98.39} & \baseline{95.58} & \baseline{93.17} & \baseline{93.32}\\
0.5 & 93.50 & \textbf{98.79} & 95.98 & 97.19 & 90.73\\
0.7 & 93.12 & 98.39 & 95.98 & 96.38 & 90.60\\
0.9 & 93.26 & 98.39 & \textbf{96.39} & \textbf{98.38} & 90.74\\ 
\midrule
$\mathcal{M}_4$ & 91.66 & 97.59 & 94.38 & 96.39 & 91.30\\
\bottomrule
\end{tabular}
\end{center}
\label{tab:sota}
\vspace{-1em}
\end{table}

\paragraph{Weight of the Learning-based Score}.
As demonstrated in Equation \ref{eq:final_score}, the final score is derived from the combined weighted contributions of the rule-based and the learning-based scores. This study examines the impact of the weight parameter $\alpha$, with the findings detailed in Table \ref{tab:abl_alpha}. Firstly, it is observed that incorporating the learning-based score significantly enhances performance compared to relying solely on the rule-based score (\ie, $\alpha=0$). The limitation of the rule-based score lies in its hand-crafted nature, which may not accurately represent all possible scenarios, whereas the learning-based score offers greater generalizability by dynamically adapting to the input features. Furthermore, a combined approach proves superior to using a purely learning-based score ($\mathcal{M}_4$), indicating that the current model, while advanced, still benefits from the inclusion of a rule-based component as a form of safety assurance. Based on optimal performance, $\alpha=0.3$ has been selected as the default setting.

\begin{table}[t]
\begin{center}
\caption{Constant velocity vs.\ learned prediction (Based on $\mathcal{M}_5$)}
\label{tab:abl_pred}
\vspace{-0.5em}
\setlength{\tabcolsep}{7pt}
\renewcommand{\arraystretch}{1.2}
\small
\begin{tabular}{y{40}|x{25}x{25}x{25}x{25}x{25}}
\toprule
Prediction & Score & Collisions & TTC & Comfort & Progress \\
\midrule
Const. vel. & 92.82 & 97.79 & 95.19 & 90.76 & 93.18\\
\baseline{Learned} & \baseline{93.57} & \baseline{98.39} & \baseline{95.58} & \baseline{93.17} & \baseline{93.32}\\
\bottomrule
\end{tabular}
\end{center}
\label{tab:sota}
\end{table}

\paragraph{Prediction method}. 
In this study, we contrast the performance of our learned prediction model against the constant velocity prediction employed in PDM-Closed and presented in Table \ref{tab:abl_pred}. It is evident that employing learned predictions for planning yields superior results compared to the simplistic constant velocity prediction, as it can more accurately discern the behaviors of the agents. Despite our prediction model producing only a singular modal trajectory for each agent, it works well in practice. 


\section{Conclusion}


In this study, we introduce \pluto{}, a pioneering data-driven planning framework that extends the capabilities of imitation learning within the autonomous driving domain. We propose innovative solutions concerning model architecture, data augmentation, and the learning framework, effectively addressing enduring challenges in imitation learning. The query-based model architecture furnishes the planner with the capacity for adaptable driving behaviors across both longitudinal and lateral dimensions. Our novel method for computing auxiliary loss, based on differentiable interpolation, offers a new approach for integrating constraints into the model. Additionally, the employment of a contrastive imitation learning framework, coupled with an advanced set of data augmentation techniques, enhances the acquisition of desired behaviors and comprehension of intrinsic interactions. Experimental evaluations utilizing real-world driving datasets demonstrate that our approach sets a new benchmark for closed-loop performance in the field. Notably, \pluto{} surpasses the previously best-performing rule-based planner, establishing a significant breakthrough in autonomous driving research.

\paragraph{Limitations and Future Work}.
In our approach, we predict a single trajectory for each dynamic agent. This methodology yields satisfactory outcomes in practical applications; nevertheless, the generation of meaningful joint multimodal predictions and their efficient incorporation into planning strategies represent significant areas for future research. The addition of a post-processing module has been demonstrated to improve overall performance effectively. However, it cannot handle scenarios where all generated trajectories are unusable. Transitioning the post-processing function to an intermediary role that directly influences trajectory generation could present a more advantageous strategy.

 




\bibliographystyle{IEEEtran}
\bibliography{IEEEabrv,ref}

\begin{thebibliography}{10}
\providecommand{\url}[1]{#1}
\csname url@samestyle\endcsname
\providecommand{\newblock}{\relax}
\providecommand{\bibinfo}[2]{#2}
\providecommand{\BIBentrySTDinterwordspacing}{\spaceskip=0pt\relax}
\providecommand{\BIBentryALTinterwordstretchfactor}{4}
\providecommand{\BIBentryALTinterwordspacing}{\spaceskip=\fontdimen2\font plus
\BIBentryALTinterwordstretchfactor\fontdimen3\font minus \fontdimen4\font\relax}
\providecommand{\BIBforeignlanguage}[2]{{%
\expandafter\ifx\csname l@#1\endcsname\relax
\typeout{** WARNING: IEEEtran.bst: No hyphenation pattern has been}%
\typeout{** loaded for the language `#1'. Using the pattern for}%
\typeout{** the default language instead.}%
\else
\language=\csname l@#1\endcsname
\fi
#2}}
\providecommand{\BIBdecl}{\relax}
\BIBdecl

\bibitem{teng2023motion}
S.~Teng, X.~Hu, P.~Deng, B.~Li, Y.~Li, Y.~Ai, D.~Yang, L.~Li, Z.~Xuanyuan, F.~Zhu \emph{et~al.}, ``Motion planning for autonomous driving: The state of the art and future perspectives,'' \emph{IEEE Transactions on Intelligent Vehicles}, 2023.

\bibitem{Dauner2023CORL}
D.~Dauner, M.~Hallgarten, A.~Geiger, and K.~Chitta, ``Parting with misconceptions about learning-based vehicle motion planning,'' in \emph{Conference on Robot Learning (CoRL)}, 2023.

\bibitem{cheng2023plantf}
J.~Cheng, Y.~Chen, X.~Mei, B.~Yang, B.~Li, and M.~Liu, ``Rethinking imitation-based planner for autonomous driving,'' in \emph{International Conference on Robotics and Automation (ICRA)}, 2024.

\bibitem{cheng2022mpnp}
J.~Cheng, R.~Xin, S.~Wang, and M.~Liu, ``Mpnp: Multi-policy neural planner for urban driving,'' in \emph{2022 IEEE/RSJ International Conference on Intelligent Robots and Systems (IROS)}.\hskip 1em plus 0.5em minus 0.4em\relax IEEE, 2022, pp. 10\,549--10\,554.

\bibitem{bansal2018chauffeurnet}
M.~Bansal, A.~Krizhevsky, and A.~Ogale, ``Chauffeurnet: Learning to drive by imitating the best and synthesizing the worst,'' \emph{arXiv preprint arXiv:1812.03079}, 2018.

\bibitem{muller2005offroad}
U.~Muller, J.~Ben, E.~Cosatto, B.~Flepp, and Y.~Cun, ``Off-road obstacle avoidance through end-to-end learning,'' \emph{Advances in neural information processing systems}, vol.~18, 2005.

\bibitem{wang2019monocular}
D.~Wang, C.~Devin, Q.-Z. Cai, P.~Kr{\"a}henb{\"u}hl, and T.~Darrell, ``Monocular plan view networks for autonomous driving,'' in \emph{2019 IEEE/RSJ International Conference on Intelligent Robots and Systems (IROS)}.\hskip 1em plus 0.5em minus 0.4em\relax IEEE, 2019, pp. 2876--2883.

\bibitem{wen2020fighting}
C.~Wen, J.~Lin, T.~Darrell, D.~Jayaraman, and Y.~Gao, ``Fighting copycat agents in behavioral cloning from observation histories,'' \emph{Advances in Neural Information Processing Systems}, vol.~33, pp. 2564--2575, 2020.

\bibitem{zhou2021exploring}
J.~Zhou, R.~Wang, X.~Liu, Y.~Jiang, S.~Jiang, J.~Tao, J.~Miao, and S.~Song, ``Exploring imitation learning for autonomous driving with feedback synthesizer and differentiable rasterization,'' in \emph{2021 IEEE/RSJ International Conference on Intelligent Robots and Systems (IROS)}.\hskip 1em plus 0.5em minus 0.4em\relax IEEE, 2021, pp. 1450--1457.

\bibitem{de2019causal}
P.~De~Haan, D.~Jayaraman, and S.~Levine, ``Causal confusion in imitation learning,'' \emph{Advances in neural information processing systems}, vol.~32, 2019.

\bibitem{cultrera2023addressing}
L.~Cultrera, F.~Becattini, L.~Seidenari, P.~Pala, and A.~Del~Bimbo, ``Addressing limitations of state-aware imitation learning for autonomous driving,'' \emph{IEEE Transactions on Intelligent Vehicles}, 2023.

\bibitem{lu2022imitation_not_enough}
Y.~Lu, J.~Fu, G.~Tucker, X.~Pan, E.~Bronstein, B.~Roelofs, B.~Sapp, B.~White, A.~Faust, S.~Whiteson \emph{et~al.}, ``Imitation is not enough: Robustifying imitation with reinforcement learning for challenging driving scenarios,'' \emph{NeurIPS 2022 Workshop on Machine Learning for Autonomous Driving}, 2022.

\bibitem{chen2020simple}
T.~Chen, S.~Kornblith, M.~Norouzi, and G.~Hinton, ``A simple framework for contrastive learning of visual representations,'' in \emph{International conference on machine learning}.\hskip 1em plus 0.5em minus 0.4em\relax PMLR, 2020, pp. 1597--1607.

\bibitem{caesar2021nuplan}
H.~Caesar, J.~Kabzan, K.~S. Tan, W.~K. Fong, E.~Wolff, A.~Lang, L.~Fletcher, O.~Beijbom, and S.~Omari, ``nuplan: A closed-loop ml-based planning benchmark for autonomous vehicles,'' \emph{Proc. IEEE Conf. on Computer Vision and Pattern Recognition (CVPR) Workshops}, 2021.

\bibitem{chen2023e2e}
L.~Chen, P.~Wu, K.~Chitta, B.~Jaeger, A.~Geiger, and H.~Li, ``End-to-end autonomous driving: Challenges and frontiers,'' \emph{arXiv preprint arXiv:2306.16927}, 2023.

\bibitem{chen2020lbc}
D.~Chen, B.~Zhou, V.~Koltun, and P.~Kr{\"a}henb{\"u}hl, ``Learning by cheating,'' in \emph{Conference on Robot Learning}.\hskip 1em plus 0.5em minus 0.4em\relax PMLR, 2020, pp. 66--75.

\bibitem{codevilla2019cilrs}
F.~Codevilla, E.~Santana, A.~M. L{\'o}pez, and A.~Gaidon, ``Exploring the limitations of behavior cloning for autonomous driving,'' in \emph{Proceedings of the IEEE/CVF International Conference on Computer Vision}, 2019, pp. 9329--9338.

\bibitem{chitta2022transfuser}
K.~Chitta, A.~Prakash, B.~Jaeger, Z.~Yu, K.~Renz, and A.~Geiger, ``Transfuser: Imitation with transformer-based sensor fusion for autonomous driving,'' \emph{IEEE Transactions on Pattern Analysis and Machine Intelligence}, 2022.

\bibitem{chitta2021neat}
K.~Chitta, A.~Prakash, and A.~Geiger, ``Neat: Neural attention fields for end-to-end autonomous driving,'' in \emph{Proceedings of the IEEE/CVF International Conference on Computer Vision}, 2021, pp. 15\,793--15\,803.

\bibitem{shao2023interfuser}
H.~Shao, L.~Wang, R.~Chen, H.~Li, and Y.~Liu, ``Safety-enhanced autonomous driving using interpretable sensor fusion transformer,'' in \emph{Conference on Robot Learning}.\hskip 1em plus 0.5em minus 0.4em\relax PMLR, 2023, pp. 726--737.

\bibitem{jia2023think}
X.~Jia, P.~Wu, L.~Chen, J.~Xie, C.~He, J.~Yan, and H.~Li, ``Think twice before driving: Towards scalable decoders for end-to-end autonomous driving,'' in \emph{Proceedings of the IEEE/CVF Conference on Computer Vision and Pattern Recognition}, 2023, pp. 21\,983--21\,994.

\bibitem{chen2022lav}
D.~Chen and P.~Kr{\"a}henb{\"u}hl, ``Learning from all vehicles,'' in \emph{Proceedings of the IEEE/CVF Conference on Computer Vision and Pattern Recognition}, 2022, pp. 17\,222--17\,231.

\bibitem{hu2023planning}
Y.~Hu, J.~Yang, L.~Chen, K.~Li, C.~Sima, X.~Zhu, S.~Chai, S.~Du, T.~Lin, W.~Wang \emph{et~al.}, ``Planning-oriented autonomous driving,'' in \emph{Proceedings of the IEEE/CVF Conference on Computer Vision and Pattern Recognition}, 2023, pp. 17\,853--17\,862.

\bibitem{jiang2023vad}
B.~Jiang, S.~Chen, Q.~Xu, B.~Liao, J.~Chen, H.~Zhou, Q.~Zhang, W.~Liu, C.~Huang, and X.~Wang, ``Vad: Vectorized scene representation for efficient autonomous driving,'' \emph{arXiv preprint arXiv:2303.12077}, 2023.

\bibitem{hu2022st-p3}
S.~Hu, L.~Chen, P.~Wu, H.~Li, J.~Yan, and D.~Tao, ``St-p3: End-to-end vision-based autonomous driving via spatial-temporal feature learning,'' in \emph{European Conference on Computer Vision}.\hskip 1em plus 0.5em minus 0.4em\relax Springer, 2022, pp. 533--549.

\bibitem{dosovitskiy2017carla}
A.~Dosovitskiy, G.~Ros, F.~Codevilla, A.~Lopez, and V.~Koltun, ``Carla: An open urban driving simulator,'' in \emph{Conference on robot learning}.\hskip 1em plus 0.5em minus 0.4em\relax PMLR, 2017, pp. 1--16.

\bibitem{vitelli2022safetynet}
M.~Vitelli, Y.~Chang, Y.~Ye, A.~Ferreira, M.~Wo{\l}czyk, B.~Osi{\'n}ski, M.~Niendorf, H.~Grimmett, Q.~Huang, A.~Jain \emph{et~al.}, ``Safetynet: Safe planning for real-world self-driving vehicles using machine-learned policies,'' in \emph{2022 International Conference on Robotics and Automation (ICRA)}.\hskip 1em plus 0.5em minus 0.4em\relax IEEE, 2022, pp. 897--904.

\bibitem{scheel2022urban}
O.~Scheel, L.~Bergamini, M.~Wolczyk, B.~Osi{\'n}ski, and P.~Ondruska, ``Urban driver: Learning to drive from real-world demonstrations using policy gradients,'' in \emph{Conference on Robot Learning}.\hskip 1em plus 0.5em minus 0.4em\relax PMLR, 2022, pp. 718--728.

\bibitem{pini2023safepathnet}
S.~Pini, C.~S. Perone, A.~Ahuja, A.~S.~R. Ferreira, M.~Niendorf, and S.~Zagoruyko, ``Safe real-world autonomous driving by learning to predict and plan with a mixture of experts,'' in \emph{2023 IEEE International Conference on Robotics and Automation (ICRA)}.\hskip 1em plus 0.5em minus 0.4em\relax IEEE, 2023, pp. 10\,069--10\,075.

\bibitem{huang2023dipp}
Z.~Huang, H.~Liu, J.~Wu, and C.~Lv, ``Differentiable integrated motion prediction and planning with learnable cost function for autonomous driving,'' \emph{IEEE transactions on neural networks and learning systems}, 2023.

\bibitem{huang2023gameformer}
Z.~Huang, H.~Liu, and C.~Lv, ``Gameformer: Game-theoretic modeling and learning of transformer-based interactive prediction and planning for autonomous driving,'' in \emph{Proceedings of the IEEE/CVF International Conference on Computer Vision}, 2023, pp. 3903--3913.

\bibitem{ross2011reduction}
S.~Ross, G.~Gordon, and D.~Bagnell, ``A reduction of imitation learning and structured prediction to no-regret online learning,'' in \emph{Proceedings of the fourteenth international conference on artificial intelligence and statistics}.\hskip 1em plus 0.5em minus 0.4em\relax JMLR Workshop and Conference Proceedings, 2011, pp. 627--635.

\bibitem{hadsell2006dimensionality}
R.~Hadsell, S.~Chopra, and Y.~LeCun, ``Dimensionality reduction by learning an invariant mapping,'' in \emph{2006 IEEE computer society conference on computer vision and pattern recognition (CVPR'06)}, vol.~2.\hskip 1em plus 0.5em minus 0.4em\relax IEEE, 2006, pp. 1735--1742.

\bibitem{he2020momentum}
K.~He, H.~Fan, Y.~Wu, S.~Xie, and R.~Girshick, ``Momentum contrast for unsupervised visual representation learning,'' in \emph{Proceedings of the IEEE/CVF conference on computer vision and pattern recognition}, 2020, pp. 9729--9738.

\bibitem{radford2021learning}
A.~Radford, J.~W. Kim, C.~Hallacy, A.~Ramesh, G.~Goh, S.~Agarwal, G.~Sastry, A.~Askell, P.~Mishkin, J.~Clark \emph{et~al.}, ``Learning transferable visual models from natural language supervision,'' in \emph{International conference on machine learning}.\hskip 1em plus 0.5em minus 0.4em\relax PMLR, 2021, pp. 8748--8763.

\bibitem{liu2021social}
Y.~Liu, Q.~Yan, and A.~Alahi, ``Social nce: Contrastive learning of socially-aware motion representations,'' in \emph{Proceedings of the IEEE/CVF International Conference on Computer Vision}, 2021, pp. 15\,118--15\,129.

\bibitem{halawa2022action}
M.~Halawa, O.~Hellwich, and P.~Bideau, ``Action-based contrastive learning for trajectory prediction,'' in \emph{European Conference on Computer Vision}.\hskip 1em plus 0.5em minus 0.4em\relax Springer, 2022, pp. 143--159.

\bibitem{wang2023fend}
Y.~Wang, P.~Zhang, L.~Bai, and J.~Xue, ``Fend: A future enhanced distribution-aware contrastive learning framework for long-tail trajectory prediction,'' in \emph{Proceedings of the IEEE/CVF Conference on Computer Vision and Pattern Recognition}, 2023, pp. 1400--1409.

\bibitem{cheng2023forecast}
J.~Cheng, X.~Mei, and M.~Liu, ``{Forecast-MAE}: Self-supervised pre-training for motion forecasting with masked autoencoders,'' \emph{Proceedings of the IEEE/CVF International Conference on Computer Vision}, 2023.

\bibitem{qi2017pointnet}
C.~R. Qi, H.~Su, K.~Mo, and L.~J. Guibas, ``Pointnet: Deep learning on point sets for 3d classification and segmentation,'' in \emph{Proceedings of the IEEE conference on computer vision and pattern recognition}, 2017, pp. 652--660.

\bibitem{zhou2023query}
Z.~Zhou, J.~Wang, Y.-H. Li, and Y.-K. Huang, ``Query-centric trajectory prediction,'' in \emph{Proceedings of the IEEE/CVF Conference on Computer Vision and Pattern Recognition}, 2023, pp. 17\,863--17\,873.

\bibitem{vaswani2017attention}
A.~Vaswani, N.~Shazeer, N.~Parmar, J.~Uszkoreit, L.~Jones, A.~N. Gomez, {\L}.~Kaiser, and I.~Polosukhin, ``Attention is all you need,'' \emph{Advances in neural information processing systems}, vol.~30, 2017.

\bibitem{carion2020end}
N.~Carion, F.~Massa, G.~Synnaeve, N.~Usunier, A.~Kirillov, and S.~Zagoruyko, ``End-to-end object detection with transformers,'' in \emph{European conference on computer vision}.\hskip 1em plus 0.5em minus 0.4em\relax Springer, 2020, pp. 213--229.

\bibitem{liu2021multimodal}
Y.~Liu, J.~Zhang, L.~Fang, Q.~Jiang, and B.~Zhou, ``Multimodal motion prediction with stacked transformers,'' in \emph{Proceedings of the IEEE/CVF Conference on Computer Vision and Pattern Recognition}, 2021, pp. 7577--7586.

\bibitem{ngiam2021scene}
J.~Ngiam, B.~Caine, V.~Vasudevan, Z.~Zhang, H.-T.~L. Chiang, J.~Ling, R.~Roelofs, A.~Bewley, C.~Liu, A.~Venugopal \emph{et~al.}, ``Scene transformer: A unified architecture for predicting multiple agent trajectories,'' \emph{arXiv preprint arXiv:2106.08417}, 2021.

\bibitem{williams1989learning}
R.~J. Williams and D.~Zipser, ``A learning algorithm for continually running fully recurrent neural networks,'' \emph{Neural computation}, vol.~1, no.~2, pp. 270--280, 1989.

\bibitem{girshick2015fast}
R.~Girshick, ``Fast r-cnn,'' in \emph{Proceedings of the IEEE international conference on computer vision}, 2015, pp. 1440--1448.

\bibitem{shi2022motion}
S.~Shi, L.~Jiang, D.~Dai, and B.~Schiele, ``Motion transformer with global intention localization and local movement refinement,'' \emph{Advances in Neural Information Processing Systems}, vol.~35, pp. 6531--6543, 2022.

\bibitem{cheng2022gpir}
J.~Cheng, Y.~Chen, Q.~Zhang, L.~Gan, C.~Liu, and M.~Liu, ``Real-time trajectory planning for autonomous driving with gaussian process and incremental refinement,'' in \emph{2022 International Conference on Robotics and Automation (ICRA)}.\hskip 1em plus 0.5em minus 0.4em\relax IEEE, 2022, pp. 8999--9005.

\bibitem{goldberger2004neighbourhood}
J.~Goldberger, G.~E. Hinton, S.~Roweis, and R.~R. Salakhutdinov, ``Neighbourhood components analysis,'' \emph{Advances in neural information processing systems}, vol.~17, 2004.

\bibitem{treiber2000idm}
M.~Treiber, A.~Hennecke, and D.~Helbing, ``Congested traffic states in empirical observations and microscopic simulations,'' \emph{Physical review E}, vol.~62, no.~2, p. 1805, 2000.

\bibitem{nuplan2023metrics}
\BIBentryALTinterwordspacing
Motional. nuplan metrics. [Online]. Available: \url{https://nuplan-devkit.readthedocs.io/en/latest/metrics_description.html}
\BIBentrySTDinterwordspacing

\bibitem{hallgarten2023gc-pgp}
M.~Hallgarten, M.~Stoll, and A.~Zell, ``From prediction to planning with goal conditioned lane graph traversals,'' \emph{arXiv preprint arXiv:2302.07753}, 2023.

\end{thebibliography}

\begin{IEEEbiography}[{\includegraphics[width=1.0in,height=1.2in,clip,keepaspectratio]{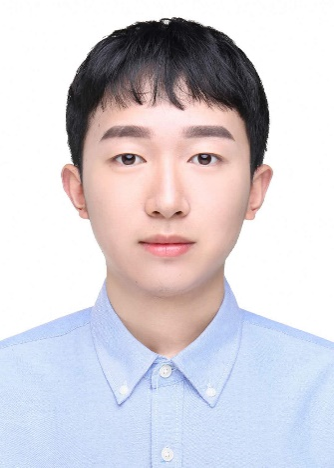}}]{Jie Cheng}
 received the B.S. degree from Huazhong University of Science and Technology, Wuhan, China, in 2019. 
 He is currently pursuing the Ph.D. degree in the Department of Electronic and Computer Engineering, the Hong Kong University of Science and Technology, HKSAR, China, supervised by Prof. Qifeng Chen.
 His research mainly focuses on motion planning and motion forecasting for autonomous driving.
\end{IEEEbiography}
\vspace{-25pt}

\begin{IEEEbiography}[{\includegraphics[width=1.0in,height=1.2in,clip,keepaspectratio]{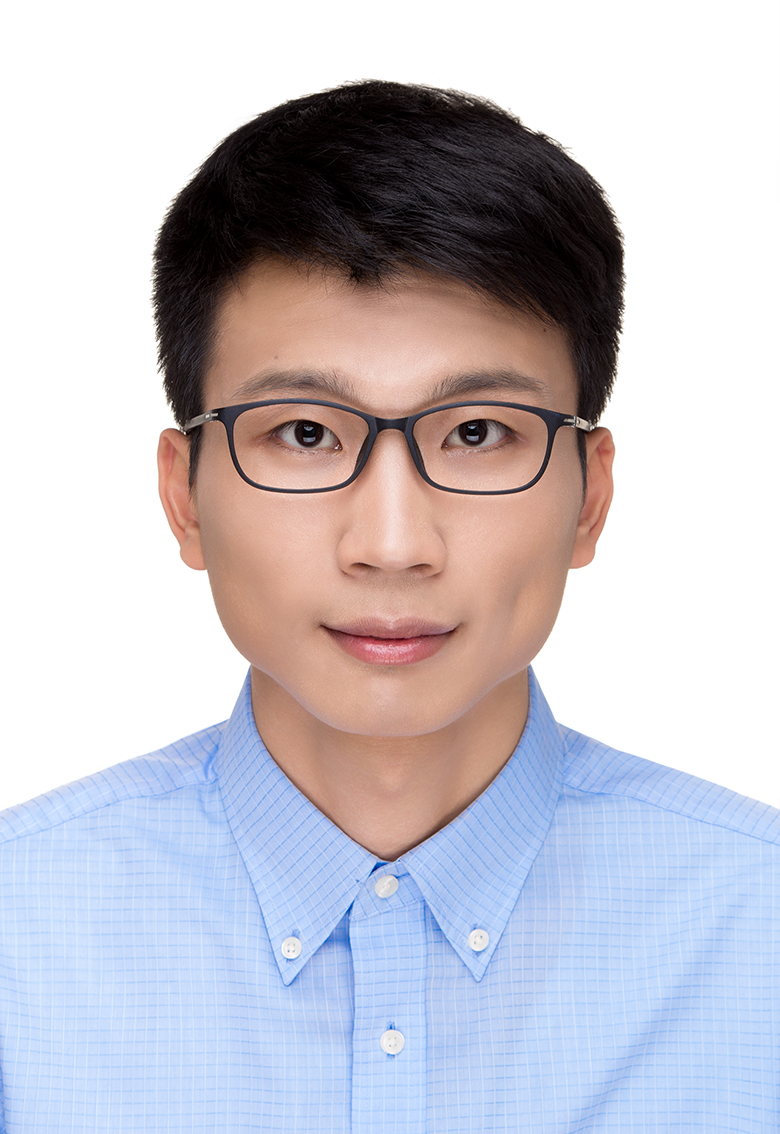}}]{Yingbing Chen}
(Student Member, IEEE) received the B.Sc. degree from Northwestern Polytechnical University, Xian, China, in 2015, and the the M.S. degree from Xiamen University, Xiamen, China, in 2018.  He is currently working toward the Ph.D. degree with the Robotics and MultiPerception Laboratory, Hong Kong University of Science and Technology, Hong Kong. His current research interests include machine learning, motion and behavioral planning for autonomous driving and robotics.
\end{IEEEbiography}
\vspace{-25pt}

\begin{IEEEbiography}[{\includegraphics[width=1.0in,height=1.2in,clip,keepaspectratio]{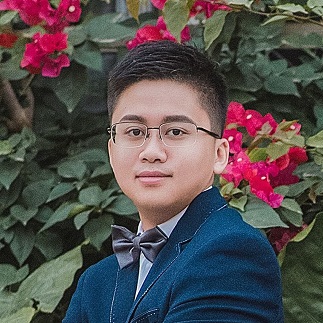}}]{Qifeng Chen}
is an assistant professor of the Department of Computer Science and Engineering and the Department of Electronic and Computer Engineering at The Hong Kong University of Science and Technology. He received his Ph.D. in computer science from Stanford University in 2017. His research interests include image 
processing and synthesis and 3D vision. He won the MIT Tech Review’s 35 Innovators under 35 in China and the Google Faculty Research Award in 2018.
\end{IEEEbiography}

\vfill

\end{document}